%% file: egpaper_for_review.tex
\documentclass[10pt,twocolumn,letterpaper]{article}

\usepackage{iccv}
\usepackage{times}
\usepackage{epsfig}
\usepackage{graphicx}
\usepackage{amsmath}
\usepackage{amssymb}

\usepackage{graphicx}
\usepackage{amsmath}
\usepackage{amssymb}
\usepackage{booktabs}
\usepackage{mathtools}

\usepackage[table]{xcolor}
\usepackage{hhline}
\usepackage[normalem]{ulem}
\usepackage{color}

\usepackage[pagebackref=true,breaklinks=true,letterpaper=true,colorlinks,bookmarks=false]{hyperref}

\usepackage[capitalize]{cleveref}

\crefname{section}{Sec.}{Secs.}
\Crefname{section}{Section}{Sections}
\Crefname{table}{Table}{Tables}
\crefname{table}{Tab.}{Tabs.}
\iccvfinalcopy 

\def\iccvPaperID{10388} 

\ificcvfinal\fi

\newcommand{\name}{\textit{FreqMRN}}

\definecolor{gray}{rgb}{0.85,0.85,0.85}

\input{math_commands}

\begin{document}

\title{Towards Accurate Human Motion Prediction via Iterative Refinement}

\author{Jiarui Sun, Girish Chowdhary\\
Dept. of Electrical and Computer Engineering\\
University of Illinois at Urbana, Champaign, USA\\
{\tt\small \{jsun57, girishc\}@illinois.edu}
}

\maketitle
\ificcvfinal\thispagestyle{empty}\fi

\begin{abstract}
   Human motion prediction aims to forecast an upcoming pose sequence given a past human motion trajectory.
   To address the problem, in this work we propose \name, a human motion prediction framework that takes into account both the kinematic structure of the human body and the temporal smoothness nature of motion.
   Specifically, \name~first generates a fixed-size motion history summary using a motion attention module, which helps avoid inaccurate motion predictions due to excessively long motion inputs.
   Then, supervised by the proposed spatial-temporal-aware, velocity-aware and global-smoothness-aware losses, \name~iteratively refines the predicted motion though the proposed motion refinement module, which converts motion representations back and forth between pose space and frequency space.
   We evaluate \name~on several standard benchmark datasets, including Human3.6M, AMASS and 3DPW.
   Experimental results demonstrate that \name~outperforms previous methods by large margins for both short-term and long-term predictions, while demonstrating superior robustness.
\end{abstract}

\input{files/introduction}

\input{files/related_work}

\input{files/method}
\input{files/experiments}
\input{files/conclusion}
{\small
\bibliographystyle{ieee_fullname}
\bibliography{egpaper_for_review}
}
\input{files/appendix}

\end{document}

%% file: math_commands.tex
\usepackage{amsmath,amsfonts,bm}

\def\eqref#1{equation~\ref{#1}}

\def\1{\bm{1}}

\def\vc{{\bm{c}}}

\def\vk{{\bm{k}}}

\def\vq{{\bm{q}}}

\def\vx{{\bm{x}}}

\def\mA{{\bm{A}}}

\def\mG{{\bm{G}}}

\def\mS{{\bm{S}}}

\def\mV{{\bm{V}}}
\def\mW{{\bm{W}}}
\def\mX{{\bm{X}}}

\DeclareMathAlphabet{\mathsfit}{\encodingdefault}{\sfdefault}{m}{sl}
\SetMathAlphabet{\mathsfit}{bold}{\encodingdefault}{\sfdefault}{bx}{n}

\def\gD{{\mathcal{D}}}

\def\gF{{\mathcal{F}}}

\def\gL{{\mathcal{L}}}

\def\sR{{\mathbb{R}}}



%% file: files/introduction.tex
\section{Introduction}

Human motion prediction aims to forecast an upcoming pose sequence given a past human motion trajectory.
This task is essential for various applications, such as human tracking \cite{DBLP:conf/cvpr/UrtasunFF06}, motion generation \cite{DBLP:journals/tog/KovarGP02}, robotics \cite{DBLP:conf/iros/GuiZWLMV18}, and autonomous driving \cite{DBLP:journals/tiv/PadenCYYF16}. 
This is a challenging task due to the large degrees of freedom of the human body, and the variability of human motion.
Early data-driven methods relied on Markov models \cite{DBLP:conf/cvpr/LehrmannGN14}, Gaussian processes \cite{DBLP:journals/pami/WangFH08},  or binary latent variables \cite{DBLP:conf/nips/TaylorHR06}. 
These methods capture the underlying dynamics of simple human activities, such as walking and golf swinging. 
However, these methods are inadequate to model more complicated motions such as playing basketball and walking a dog.

Modern methods aim to improve forecast of complicated human motion using large-scale datasets and deep learning.
Among these methods, some utilize Recurrent Neural Networks (RNNs) to model temporal dynamics due to the sequential nature of human motion \cite{DBLP:conf/iccv/FragkiadakiLFM15, DBLP:conf/cvpr/JainZSS16, DBLP:conf/cvpr/MartinezB017}. 
However, besides the well-known problem of training difficulty, RNN methods suffer greatly from error accumulation, which can lead to unrealistic predicted motion as the forecasting horizon increases. 
Recent work has shown that feed-forward methods instead can result in much better performance.
Some such works use convolutional neural networks (CNNs) to learn structural and temporal correlations by treating pose sequences from human motion trajectories as regular grid structure data \cite{DBLP:conf/cvpr/ButepageBKK17, DBLP:conf/cvpr/LiZLL18}. 
Other works employ the attention mechanism, which enables models to capture dependencies on arbitrary structures and temporal scales \cite{DBLP:conf/3dim/AksanKCH21, DBLP:conf/eccv/MaoLS20, DBLP:conf/cvpr/MedjaouriD22}. 
There are also works focusing on the simplicity aspect, designing lightweight and effective models based on multi-layer perceptrons (MLPs) \cite{DBLP:conf/ijcai/BouaziziHKDB22, DBLP:conf/wacv/GuoDSLAM23}. 
These feed-forward approaches have not only been pushing the boundary of prediction accuracy but also greatly improve model efficiency compared to RNN-based methods.

\begin{figure*}[!tb]
\centering
\includegraphics[width=0.7\linewidth]{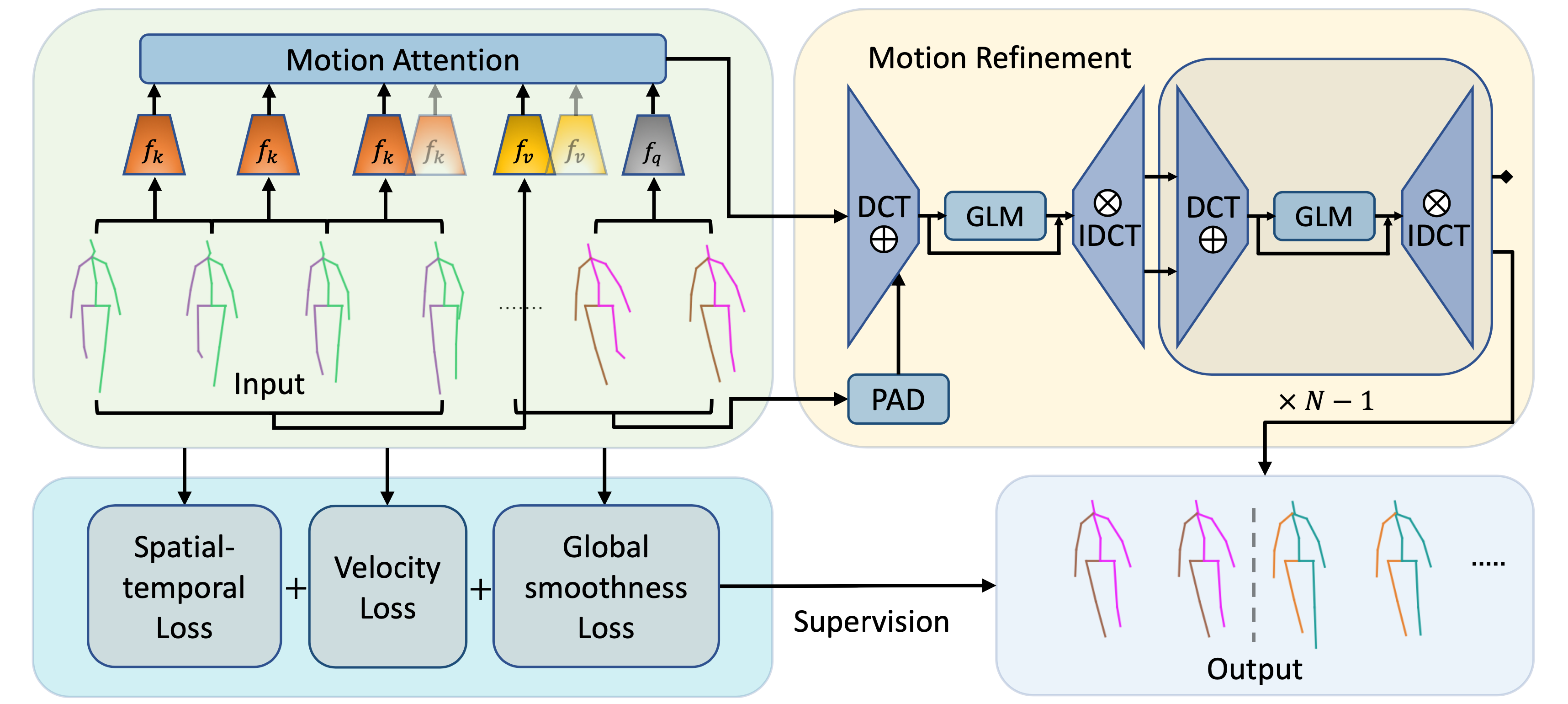}
\caption{The proposed \name~framework. \name~first generates motion summary through the motion attention module, then uses the motion refinement module to predict the entire motion sequence. The model is supervised by the loss that is spatial-temporal-aware, velocity-aware and global-smoothness-aware.}
\label{fig:mdl}
\end{figure*}
Among feed-forward approaches, the Discrete Cosine Transform (DCT) based methods stand out due to their superior performance \cite{DBLP:conf/iccv/DangNLZL21, DBLP:conf/eccv/MaoLS20, DBLP:conf/iccv/MaoLSL19, DBLP:journals/ijcv/MaoLSL21}. 
Instead of learning the mapping from past motion to future motion in the pose space directly, these methods learn the differences between the padded, transformed past pose sequence and the transformed entire pose sequence in the frequency space. 
Specifically, they first pad the last observed pose repetitively onto the past motion trajectory until the extended input sequence has the same length as the entire motion trajectory.
Then, by performing DCT on the extended input, these methods create frequency components that are very similar to those of the transformed entire pose sequence.
Finally, residual learning \cite{DBLP:conf/cvpr/HeZRS16} is used to make predictions.
Since the frequency differences are relatively small, this technique greatly simplifies the learning problem and thus improves prediction accuracy.
Another shared principle of these approaches is that they utilize Graph Convolutional Networks (GCNs) to learn the structural relationships between human body joints. 

In this paper, we present a framework to address some key limitations of DCT-based methods. 
Although these methods have greatly improved human motion prediction performance, several issues are evident.
First, it was found in \cite{DBLP:conf/iccv/MaoLSL19} that directly processing overly long motion history in frequency space hurts model performance because the model fails to capture small motion details, resulting in a lack of precision in the prediction.
However, for complicated motions that exhibit their full characterization only over a longer period of time, a longer input motion history is necessary.
Second, the last observed pose may not be the best candidate for initialization of future pose sequences \cite{DBLP:conf/cvpr/MaNLZL22}, especially when long-term future predictions are required, where the uncertainty in motion can be large.
In addition, despite being able to capture the implicit structural dependencies between joints, the GCNs used in current methods either treat all joints equally or are only aware of a simple structural hierarchy, ignoring the details of human kinematic structure.

Our main contribution to address these limitations is a framework, named \textit{\textbf{\underline{Freq}}uency space \textbf{\underline{M}}otion \textbf{\underline{R}}efinement \textbf{\underline{N}}etwork} (\name) for 3D human motion prediction, as illustrated in \cref{fig:mdl}.
\name~first employs a motion attention module that conducts motion dependency discovery and motion aggregation for the input pose sequence at the sub-series level. 
Operating in pose space, the motion attention module generates a fixed-size motion history summary based on the entire input pose sequence, thus avoiding the predictor from processing too much motion information directly in the frequency space.
Then, based on the latest observed motion as well as the motion history summary, a motion refinement module consisting of multiple graph learning modules is applied to gradually refine and forecast future pose sequence in a multi-stage manner.
At each stage, the motion refinement module first utilizes GCNs to learn joint dependencies in the frequency space; then, the predictions are transformed back into pose space and served as the refined input to the next graph learning module.
Lastly, the predicted pose sequence is supervised by a loss that is spatial-temporal-aware, velocity-aware and global-smoothness-aware.
Specifically, the loss aims to 1) take into account the degree of variation in motion in both space and time by taking into account the details of human body structure; 2) consider motion velocity and 3) consider the global smoothness of motion.
This enables \name~to capture both the spatial and temporal details of human motion and its global behavior, further regulating model predictions.
During testing, \name~generates motion trajectory of desired length in an auto-regressive manner, while incorporating information from the provided motion history as well as the previously predicted motion. 

Our experiments demonstrate that \name~outperforms previous approaches by significant margins and shows superior robustness.
Specifically, we perform experiments on three standard benchmark datasets, including Human3.6M \cite{DBLP:journals/pami/IonescuPOS14}, AMASS \cite{DBLP:conf/iccv/MahmoodGTPB19} and 3DPW \cite{DBLP:conf/eccv/MarcardHBRP18}. 
A detailed ablation study is conducted to further evaluate the benefits brought by each component of our proposed framework.
We summarize our key contributions as follows:
\begin{itemize}
\item A novel network that iteratively refines predicted human motion by learning and converting representations back and forth between frequency space and pose space. 
\item A novel framework \name~that is spatial-temporal-aware, velocity-aware and global-smoothness-aware for 3D human motion prediction, capable of emphasizing different parts of human body according to human kinematic structure.
\item A comprehensive set of experiments that show the effectiveness of \name~over state-of-the-art methods.
\end{itemize}

%% file: files/related_work.tex
\section{Related Work}

\subsection{RNN-based Methods}
\label{sec::rela_rnn}
After the advent and rapid development of deep learning, the problem of human motion prediction is largely addressed by neural network-based methods due to their superior effectiveness. 
Since human motion exhibits sequential nature as time series data, RNNs are widely used to capture its temporal dynamics.
Fragkiadaki \etal \cite{DBLP:conf/iccv/FragkiadakiLFM15} proposed the ERD model, a pioneer work that augments LSTM layers \cite{DBLP:journals/neco/HochreiterS97} with encoder and decoder networks to learn the temporal dynamics and future motion representations.
Jain \etal \cite{DBLP:conf/cvpr/JainZSS16} proposed the Structural-RNN model, which incorporates spatio-temporal graphs to RNNs.
Comparing to ERD, it further leverages the underlying high-level structure of human body.
Instead of predicting human poses directly as in \cite{DBLP:conf/iccv/FragkiadakiLFM15, DBLP:conf/cvpr/JainZSS16}, Martinez \etal \cite{DBLP:conf/cvpr/MartinezB017} proposed the residual sequence-to-sequence architecture, forcing RNNs to model velocities.
Recently, Li \etal \cite{DBLP:conf/cvpr/LiCZZW020} proposed the DMGNN model, which integrates graph structures to RNN units directly and models human motion in a multi-scale manner.
Comparing to traditional approaches \cite{DBLP:conf/nips/TaylorHR06, DBLP:journals/pami/WangFH08}, these methods are capable of predicting more complex motions with better performance.
However, RNN-based approaches are difficult to train due to their inherent sequential nature, which also leads to issues such as severe error accumulation and prediction discontinuity.

\subsection{Feed-Forward Approaches}
\label{sec::rela_ffd}

To address the issues exhibited by RNN-based methods, attention has been turned to feed-forward approaches, which are naturally parallelizable and thus much easier to train.
Some works utilize CNNs to capture both spatial and temporal correlations of human motion in a hierarchical manner \cite{DBLP:conf/cvpr/LiZLL18, DBLP:journals/tcsv/LiuYLDLL21}.
Although CNN-based methods are more efficient, the range of spatial and temporal correlations they capture is governed by the size of the convolutional filters, which quires laborious fine-tuning.

Starting from the seminal work of Mao \etal \cite{DBLP:conf/iccv/MaoLSL19}, the family of DCT-based approaches \cite{DBLP:conf/eccv/CaiHWC0YLYZSLLM20, DBLP:conf/iccv/DangNLZL21, DBLP:conf/eccv/MaoLS20, DBLP:conf/cvpr/GuoBAM22, DBLP:conf/iccv/MaoLSL19, DBLP:journals/ijcv/MaoLSL21} has rapidly emerged due to their promising performance.
Instead of learning representations in pose space, these methods model the temporal dependence of human motion in frequency space through decoupling motion trajectories into DCT coefficients.
Most of them \cite{DBLP:conf/iccv/DangNLZL21, DBLP:conf/eccv/MaoLS20, DBLP:conf/cvpr/GuoBAM22, DBLP:conf/iccv/MaoLSL19, DBLP:journals/ijcv/MaoLSL21} utilize GCNs for structural modeling, and the attention mechanism is used in \cite{DBLP:conf/eccv/CaiHWC0YLYZSLLM20, DBLP:conf/eccv/MaoLS20,  DBLP:conf/cvpr/GuoBAM22, DBLP:journals/ijcv/MaoLSL21} to capture long-term temporal correlations.
In particular, Mao \etal \cite{DBLP:journals/ijcv/MaoLSL21} used an ensemble of models, combining multiple DCT-based models at different scales to achieve better prediction accuracy.
Guo \etal \cite{DBLP:conf/cvpr/GuoBAM22} extended the problem to the two-person situation, using a cross-attention mechanism to exploit the historical information of two interacted individuals.
Dang \etal \cite{DBLP:conf/iccv/DangNLZL21} introduced the concept of hierarchy in their MSR-GCN model, while using intermediate supervision to increase model expressiveness.
Ma \etal \cite{DBLP:conf/cvpr/MaNLZL22} also adopted the idea of intermediate supervision.
Note that while our method and \cite{DBLP:conf/cvpr/MaNLZL22} both utilize multi-stage frameworks to progressively predict accurate motions, they aim to use an averaged, stage-number-based recursively smoothed target motions to guide predictions.
In contrast, our approach concurrently refines the generated motion summary and query motion based on the proposed multi-component loss independent of learning stage used, and is capable of predicting human motion with any desired input-output length without sacrificing performance.
Recently, MLP-based approaches \cite{DBLP:conf/ijcai/BouaziziHKDB22, DBLP:conf/wacv/GuoDSLAM23} and attention-based methods \cite{DBLP:conf/3dim/AksanKCH21, DBLP:conf/cvpr/MedjaouriD22} have been proposed, targeting at the simplicity and generalization aspects of human motion prediction frameworks.

%% file: files/method.tex
\section{Our Approach}
\label{sec::method}


Following previous works \cite{DBLP:conf/eccv/MaoLS20, DBLP:conf/wacv/GuoDSLAM23}, we are given human motion history $\mX_{1:H} = \{\vx_1, \vx_2, \ldots, \vx_{H}\}$, consisting of $H$ consecutive human poses $\vx_{i}$.
Our objective is to predict $F$ future human poses $\mX_{H+1:H+F} = \{\vx_{H+1}, \vx_{H+2}, \ldots, \vx_{H+F}\}$.
Each human pose $\vx_{i} \in \sR^{P}$ is described by $P$ parameters.
Since we are interested in predicting 3D joint positions, $P = J \times 3$ where $J$ denotes the number of joints.

\subsection{Motion Attention Module}
\label{subsec::mo_att}

Since most human activities involve repetitive motion patterns, leveraging motion sub-sequence similarities in long motion history can improve prediction quality.
To this end, we simply adapt the motion attention model of \cite{DBLP:conf/eccv/MaoLS20}, which summarizes motion history of arbitrary length into a fixed-size representation.   
By aggregating the most relevant partial motions from the history, the motion attention module exploits arbitrarily long historical information while avoiding performance degradation caused by directly processing too much information in the frequency space.

In order for the motion attention module to discover motion dependencies at the sub-series level, the input motion history $\mX_{1:H}$ is first divided into a collection of partial pose sequences $\{\mX_{i:i+L+F-1}\}_{i=1}^{H-L-F+1}$.
Each motion sub-sequence contains $L+F$ consecutive human poses, where $L$ denotes query length and $F$ is the length of the future motion that is to be predicted at once based on the previous $L$ poses.
The objective is to create a summarized motion representation of length $L+F$, by aggregating values $\mX_{i:i+L+F-1}$ based on the attention scores generated from query $\mX_{H-L+1:H}$ and keys $\mX_{i:i+L-1}$ for all $i$.

To this end, the motion attention module first extract query representation and key representations.
The query $\vq \in \sR^{d}$ and keys $\vk_i \in \sR^{d}$ are learned as:
\begin{equation}
\label{query_key}
\vq = f_q(\mX_{H-L+1:H}), \vk_i = f_k(\mX_{i:i+L-1}),
\end{equation}
where $f_q$, $f_k$: $\sR^{P\times L} \rightarrow \sR^{d}$ are two small CNNs mapping raw 3D motion representations to $d$-dimensional latent representations. 
Let $\mV_i \in \sR^{P\times (L+F)}$ denotes partial motion $\mX_{i:i+L+F-1}$. 
Then, the attention scores and motion summary $\mS \in \sR^{P\times (L+F)}$ are computed as:
\begin{equation}
\label{att_score}
a_i = \frac{\vq\vk_i^T}{\sum_{j=1}^{H-L-F+1}\vq\vk_j^T}, \mS = \sum_{i=1}^{H-L-F+1}a_i\mV_i,
\end{equation}
such that $\sum_{i}a_i = 1$. 
ReLU is used \cite{DBLP:conf/icml/NairH10} as the output layer for $f_q$ and $f_k$ to avoid negative $a_i$.

While being conceptually the same, the empirical differences between the original motion attention model and ours is that instead of using values $\mV_i$ in the frequency space as in \cite{DBLP:conf/eccv/MaoLS20}, our motion attention module operates only in pose space, reducing computational overhead.
More importantly, the motion summary $\mS$ is used together with the query $\mX_{H-L+1:H}$ as an auxiliary input to the motion refinement module to guide the refinement process.
During testing, if we want to generate longer future motion, the motion attention module is capable of incorporating new predictions by iteratively augmenting the motion history with new predictions as input.
As such, the motion attention module can utilize history information of arbitrary length, while avoiding generating excessively long motions that may impair model performance.

\subsection{Motion Refinement Module}
\label{subsec::irm}

The motion refinement module aims to predict future motion based on the latest observed motion $\mX_{H-L+1:H}$ and the motion summary $\mS$ generated from the motion attention module. 
The core of this module is to iteratively refine the predicted human motion by converting representations back and forth between the original 3D pose space and the DCT space.
Specifically, the module consists of two components: the domain conversion unit and the graph learning module. 
We present both of them as follows.

\noindent\textbf{Domain conversion unit.} Among DCT-based human motion prediction methods, most prior works \cite{DBLP:conf/eccv/MaoLS20, DBLP:conf/iccv/MaoLSL19, DBLP:journals/ijcv/MaoLSL21} aim to learn the differences between the padded, transformed past pose sequence and the transformed entire pose sequence in the DCT space.
Specifically, for a motion history of length $L$, these approaches first convert the original input pose sequence to $\mX'_0 \in \sR^{P\times(L+F)}$ in DCT space as:
\begin{equation}
\label{eq::padding}
 \mX_0 = [\mX_{H-L+1:H}; \mX_H \times F], \mX'_0 = \gD(\mX_0),
\end{equation}
where $\gD(\cdot)$ denotes the DCT operation.
Then, the output pose sequence $\mX_N \in \sR^{P\times(L+F)}$ is generated as:
\begin{equation}
\label{eq::one_step_end}
 \mX_N = \gD^{-1}\left(\gF\left(\mX'_0\right)+\mX'_0\right),
\end{equation}
where $\gD^{-1}(\cdot)$ denotes the inverse DCT (IDCT) operation, $\gF(\cdot)$ is the residual predictor, and $N$ represents the number of learning stages. 
The last $F$ poses of $\mX_N$ are evaluated against groundtruth $\mX_{H+1:H+F}$ for model performance.

From the above learning pipeline, it is clear that the learning difficulty is closely related to the actual differences between $\mX_0$ and $\mX_{H-L+1:H+F}$ \cite{DBLP:conf/iccv/DangNLZL21, DBLP:conf/cvpr/MaNLZL22}.
In extreme cases, the learning problem becomes trivial if the person stops moving at the prediction horizon. 
Inspired by this observation, instead of learning the future motion all at once, we learn to iteratively refine the predicted motion  through converting representations back and forth between pose space and DCT space via domain conversion units.
Recall that $N$ denotes the number of iterations.
Now, after \cref{eq::padding}, $\mX_N$ can be computed iteratively as:
\begin{equation}
\label{eq::o_iter_refine_1}
 \mX_n = \gD^{-1}\left(\gF_{n}\left(\mX'_{n-1}\right)+\mX'_{n-1}\right), \mX'_n = \gD(\mX_n),
\end{equation}
%
%
%
where $\gF_{n}(\cdot)$ is the $n^{th}$ residual predictor.

\begin{figure}[!tb]
\centering
\includegraphics[width=0.65\linewidth]{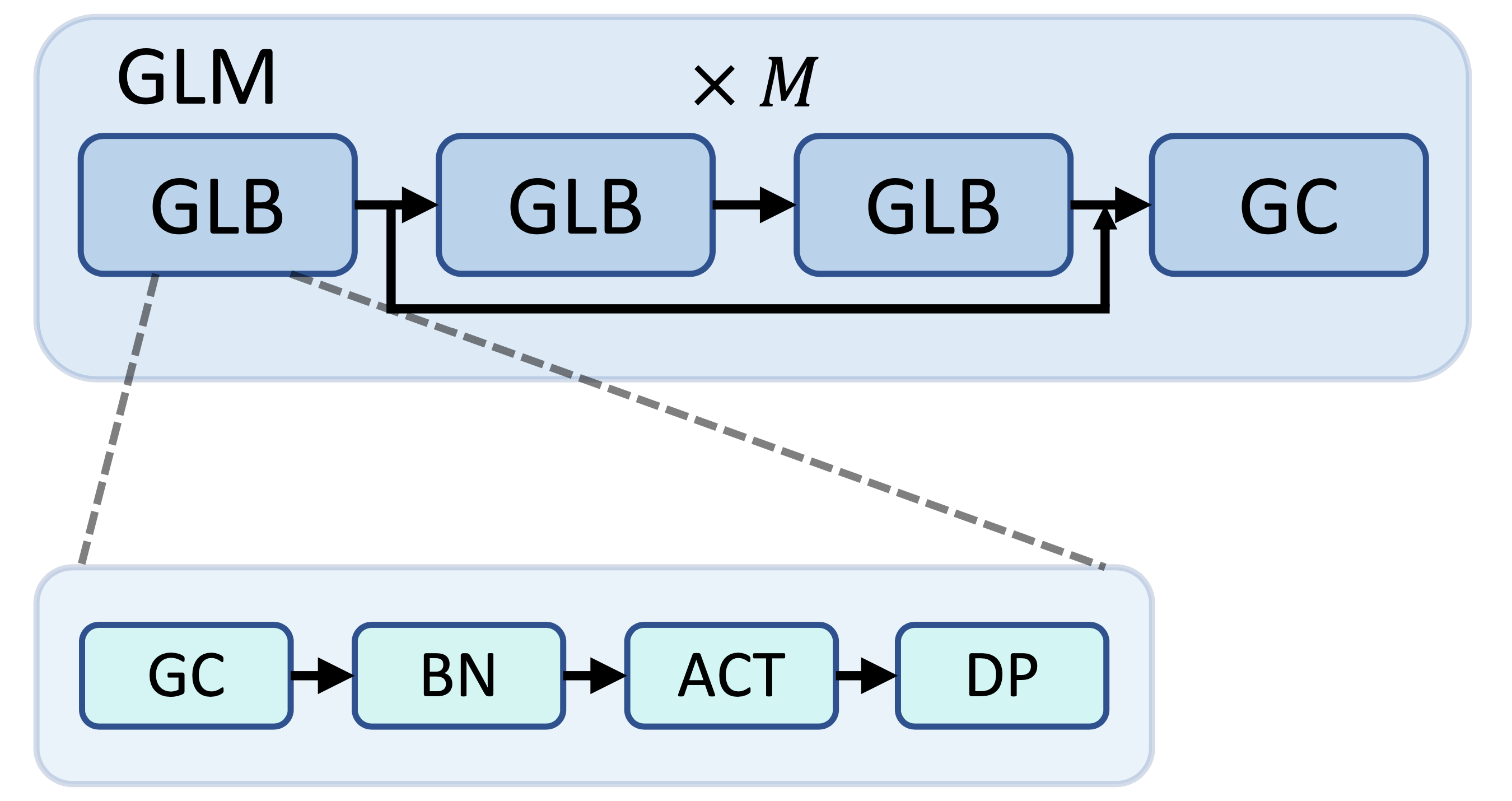}
\caption{The proposed graph learning module. It consists of multiple graph learning blocks with residual connections.}
\label{fig:glm}
\end{figure}
However, the above predictor only exploits the $L$ latest poses, as directly considering the entire $H$ past poses is detrimental to performance \cite{DBLP:conf/iccv/MaoLSL19}.
To this end, we propose to integrate the motion summary $\mS$ from the motion attention module to \cref{eq::o_iter_refine_1}. 
Let $\mS_0 = \mS$. 
The motion refinement module predicts motion as:
\begin{equation}
\label{eq::iter_refine_1}
    \mG'_{n-1} = [\gD(\mS_{n-1}); \gD(\mX_{n-1})],
\end{equation}
\begin{equation}
\label{eq::iter_refine_2}
    [\mS'_{n}; \mX'_{n}] = \gF_{n}(\mG'_{n-1}) + \mG'_{n-1},
\end{equation}
\begin{equation}
\label{eq::iter_refine_3}
    \mS_{n} = \gD^{-1}(\mS'_{n}), \mX_{n} = \gD^{-1}(\mX'_{n}),
\end{equation}
%
where $\mG'_{n-1} \in \sR^{P\times 2(L+F)}$ represents the concatenation of the motion summary and the predicted motion representation.
At the $N^{th}$ stage, $\mS_N$ is discarded.

\noindent\textbf{Graph learning module.} The residual predictor, \ie, the graph learning module (GLM) $\gF_{n}(\cdot)$ aims to capture structural and temporal dependencies across human joints and DCT components.
The architecture is shown in \cref{fig:glm}. 
It contains $K$ graph learning blocks (GLBs), each of which is based on the graph convolution (GC) operation.
At the $k^{th}$ block, the GC operation is defined as:
\begin{equation}
\label{eq::gc}
    \mathrm{GC}(\mG') = \mA_k\mG'\mW_k,
\end{equation}
where $\mA_k \in \sR^{P\times P}$ defines the learnable structural graph adjacency matrix and $\mW_k \in \sR^{d\times d'}$ denotes the weight matrix for temporal modeling at the $k^{th}$ layer.

Based on the GC operation, each GLB sequentially executes GC, batch normalization \cite{DBLP:conf/icml/IoffeS15}, tanh activation and dropout \cite{DBLP:journals/jmlr/SrivastavaHKSS14}. Let $\mG'_0 \in \sR^{P\times 2(L+F)}$ denotes the input. 
As shown in \cref{fig:glm}, we define the GLM $\gF(\cdot)$ as:
\begin{equation}
\label{eq::glm_1}
    \mG'_1 = \mathrm{GLB}_1(\mG'_0),
\end{equation}
\begin{equation}
\label{eq::glm_2}
    \mG'_{k+2} = \mathrm{GLB}_{k+2}\left(\mathrm{GLB}_{k+1}(\mG'_k)\right) + \mG'_k,
\end{equation}
\begin{equation}
\label{eq::glm_3}
    \mG'_{\mathrm{out}} = \mathrm{GC}(\mG'_{K}),
\end{equation}
where $k = 1,3,\ldots,K$ denotes GLB iteration index.
The last GC operation is used to ensure $\mG'_{\mathrm{out}} \in \sR^{P\times 2(L+F)}$, that it has the correct number of DCT components for frequency-pose conversion. 

\subsection{Learning Algorithm}
\label{subsec::learn_algo}

Using the proposed motion attention module and motion refinement module, we predict 3D human poses in an end-to-end manner.
During training, most prior works \cite{DBLP:conf/eccv/MaoLS20, DBLP:conf/wacv/GuoDSLAM23, DBLP:conf/iccv/MaoLSL19} considers the $\ell_2$ loss as supervision.
However, the $\ell_2$ loss simply treats all joints equally, ignoring this important kinematic structure of the human body. 
In addition, it also treats all poses equally across time, which leaves the error accumulation issue unaddressed.
Although such issue is not as severe as for RNNs, predicting accurate initial poses can still help predict later poses, especially for multi-stage networks.
To this end, we propose the following losses that help regulate the model to obtain accurate predictions.

\noindent\textbf{Spatial-temporal-aware loss.} To account for variations in the spatial and temporal aspects of human motion, we propose to weight the joint representations differently as:
\begin{equation}
\label{eq::st_loss_1}
\gL_{st}=\frac{1}{J(L+F)} \sum_{f=1}^{L+F} \sum_{j=1}^J\|(\mX_j^f-\tilde{\mX}_j^f) \cdot \lambda_j^f\|_2,
\end{equation}
\begin{equation}
\label{eq::st_loss_2}
\sum_{f=1}^{L+F} \sum_{j=1}^J\lambda_j^f = J(L+F),
\end{equation}
where $\mX_j^f$ and $\tilde{\mX}_j^f$ are the groundtruth and predicted representations of the $j^{th}$ joint at the $f^{th}$ frame respectively, and $\lambda_j^f$ denotes the specific assigned weight.
The weights are computed based on the specifics of the kinematic structure and their temporal index, and they do not require learning. 

Specifically, let $\mX = \{\vx_{1}, \vx_{2}, \ldots, \vx_{F}\}$ denotes predicted motion, where $\vx \in \sR^{J\times 3}$ describes the 3D coordinates of one particular human pose. 
Based on the kinematic structure of the human body, each pose $\vx$ can be described by $\mathcal{L}$ kinematic chains.
Let $\vc_l$ denote the $l^{th}$ kinematic chain, $b_l^{i}$ denote the bone length of the $i^{th}$ bone on $\vc_l$, and $l(\vc_l)$ denote the total number of bones of $\vc_l$.
Form the spatial perspective, suppose that the $j^{th}$ joint is the $j'^{th}$ joint on chain $\vc_l$.
We define:
\begin{equation}
\label{eq::lambda_s}
\lambda_j^f \propto \frac{j'}{l(\vc_l)}\ln\left(\sum_{i'=1}^{j'}b_l^{i'}\right).
\end{equation}
From the temporal perspective, suppose that the $f^{th}$ joint is the $f'^{th}$ joint enumerating from the prediction horizon.
We define: 
\begin{align}
\label{eq::lambda_t}
\lambda_j^f & \propto (F-f+L) & \text{if $f > L$,} \\
\lambda_j^f & \propto (1) & \text{otherwise.}
\end{align}
%
In particular, from the structural point of view, higher weights are assigned to joints that may exhibit vigorous motions (external joints).
From the temporal perspective, higher weights are assigned to joints that are closer to the prediction horizon.

\noindent\textbf{Velocity-aware loss.} Since most human motions tend to be smooth, following \cite{DBLP:conf/wacv/GuoDSLAM23},
we adopt the velocity-aware loss to further regulate the predicted motion trajectories.
It is computed as:
\begin{equation}
\label{eq::v_loss_1}
\gL_{v}=\frac{1}{J(L+F-1)} \sum_{f=1}^{L+F-1} \sum_{j=1}^J\|(\mV_j^f-\tilde{\mV}_j^f)\|_2,
\end{equation}
where $\mV_j^f$ and $\tilde{\mV}_j^f$ are the groundtruth and predicted joint velocities, computed as joint displacements between consecutive frames.

\noindent\textbf{Global-smoothness-aware loss.} Note that for both proposed losses, instead of only focusing on the unobserved future motion of length $F$, we also aim to reconstruct the last observed $L$ poses. 
Through reconstructing the query motion, \name~is aware of the entire pose trajectory in a global sense thus further regulate its prediction.
The learning algorithm aims to optimize the overall loss, which is defined as:
\begin{equation}
\label{eq::v_loss_total}
\gL=\gL_{st} + \gL_{v}.
\end{equation}
We provide detailed ablation analysis for the loss choices in \cref{subsec::ablation}.

%% file: files/experiments.tex
\section{Experiments}
\label{sec:exp}

In this section, we evaluate the effectiveness of \name.
We first present the datasets we used in \cref{subsec::dataset}.
Then, we introduce our experimental setup in \cref{subsec::setup}, including evaluation metrics, baseline methods for comparison and model implementation details.
The quantitative and qualitative results are presented in \cref{subsec::result}.
A detailed ablation analysis is performed and presented in \cref{subsec::ablation} to evaluate the effectiveness of the proposed framework.

\begin{table*}[t]\scriptsize
\begin{center}
\resizebox{\linewidth}{!}{
    \begin{tabular}{ |c |cccc| cccc | cccc| cccc| } 
    \hline
    scenarios  & \multicolumn{4}{c|}{walking} & \multicolumn{4}{c|}{eating} & \multicolumn{4}{c|}{smoking} & \multicolumn{4}{c|}{discussion} \\
    \hline
    milliseconds & 80ms & 400ms & 560ms & 1000ms & 80ms & 400ms & 560ms & 1000ms & 80ms & 400ms & 560ms & 1000ms & 80ms & 400ms & 560ms & 1000ms  \\ 
    \hline
    \rowcolor{gray} 
    ResSup \cite{DBLP:conf/cvpr/MartinezB017} & 23.2 & 66.1 & 71.6 & 79.1 & 16.8 & 61.7 & 74.9 & 98.0 & 18.9 & 65.4 & 78.1 & 102.1 & 25.7 & 91.3 & 109.5 & 131.8  \\ 
    convSeq2Seq \cite{DBLP:conf/cvpr/LiZLL18} & 17.7 & 63.6 & 72.2 & 82.3 & 11.0 & 48.4 & 61.3 & 87.1 & 11.6 & 48.9 & 60.0 & 81.7 & 17.1 & 77.6 & 98.1 & 129.3 \\ 
    \rowcolor{gray}
    LTD-50-25 \cite{DBLP:conf/iccv/MaoLSL19} & 12.3 & 44.4 & 50.7 & 60.3 & 7.8 & 38.6 & 51.5 & 75.8 & 8.2 & 39.5 & 50.5 & 72.1 & 11.9 & 68.1 & 88.9 & 118.5 \\
    LTD-10-10 \cite{DBLP:conf/iccv/MaoLSL19} & 11.1 & 42.9 & 53.1 & 70.7 & 7.0 & 37.3 & 51.1 & 78.6 & 7.5 & 37.5 & 49.4 & 71.8 & 10.8 & 65.8 & 88.1 & 121.6 \\
    \rowcolor{gray}
    HRI \cite{DBLP:conf/eccv/MaoLS20} & 10.0 & 39.8 & 47.4 & 58.1 & 6.4 & 36.2 & 50.0 & 75.7 & 7.0 & 36.4 & 47.6 & 69.5 & 10.2 & 65.4 & 86.6 & 119.8\\
    STDGCN \cite{DBLP:conf/cvpr/MaNLZL22} & 11.2 & 42.8 & 49.6 & 58.9 & 6.5 & 36.8 & 50.0 & 74.9 & 7.3 & 37.5 & 48.8 & 69.9 & 10.2 & 64.4 & 86.1 & $\underline{116.9}$\\
    \rowcolor{gray}
    MMA  \cite{DBLP:journals/ijcv/MaoLSL21} & 9.9 & $\underline{39.0}$ & $\underline{46.2}$ & 57.1 & 6.2 & $\underline{35.3}$ & $\underline{48.6}$ & $\underline{73.7}$ & 6.8 & $\underline{35.5}$ & $\underline{46.5}$ & $\mathbf{68.7}$ & 9.9 & $\underline{64.0}$ & $\underline{85.2}$ & 117.5 \\
    siMLPe \cite{DBLP:conf/wacv/GuoDSLAM23} & $\underline{9.9}$ & 39.6 & 46.8 & $\underline{55.7}$ & $\underline{5.9}$ & 36.1 & 49.6 & 74.5 & $\underline{6.5}$ & 36.3 & 47.2 & 69.3 & $\underline{9.4}$ & 64.3 & 85.7 & $\mathbf{116.3}$\\
    \rowcolor{gray}
    \hline
    Ours & $\mathbf{9.0}$ & $\mathbf{36.7}$ & $\mathbf{44.5}$ & $\mathbf{54.1}$ & $\mathbf{5.6}$ & $\mathbf{33.9}$ & $\mathbf{47.2}$ & $\mathbf{73.1}$ & $\mathbf{6.2}$ & $\mathbf{35.1}$ & $\mathbf{46.9}$ & $\underline{69.2}$ & $\mathbf{8.7}$ & $\mathbf{62.9}$ & $\mathbf{84.7}$ & 117.1 \\ 
    \hline
    scenarios  & \multicolumn{4}{c|}{directions} & \multicolumn{4}{c|}{greeting} & \multicolumn{4}{c|}{phoning} & \multicolumn{4}{c|}{posing} \\
    \hline
    milliseconds & 80ms & 400ms & 560ms & 1000ms & 80ms & 400ms & 560ms & 1000ms & 80ms & 400ms & 560ms & 1000ms & 80ms & 400ms & 560ms & 1000ms  \\ 
    \hline
    \rowcolor{gray}
    ResSup \cite{DBLP:conf/cvpr/MartinezB017}& 21.6 & 84.1 & 101.1 & 129.1 & 31.2 & 108.8 & 126.1 & 153.9 & 21.1 & 76.4 & 94.0 & 126.4 & 29.3 & 114.3 & 140.3 & 183.2 \\ 
    convSeq2Seq \cite{DBLP:conf/cvpr/LiZLL18}& 13.5 & 69.7 & 86.6 & 115.8 & 22.0 & 96.0 & 116.9 & 147.3 & 13.5 & 59.9 & 77.1 & 114.0 & 16.9 & 92.9 & 122.5 & 187.4 \\ 
    \rowcolor{gray}
    LTD-50-25 \cite{DBLP:conf/iccv/MaoLSL19} & 8.8 & 58.0 & 74.2 & 105.5 & 16.2 & 82.6 & 104.8 & 136.8 & 9.8 & 50.8 & 68.8 & 105.1 & 12.2 & 79.9 & 110.2 & 174.8\\
    LTD-10-10 \cite{DBLP:conf/iccv/MaoLSL19} & 8.0 & 54.9 & 76.1 & 108.8 & 14.8 & 79.7 & 104.3 & 140.2 & 9.3 & 49.7 & 68.7 & 105.1 & 10.9 & 75.9 & 109.9 & 171.7\\
    \rowcolor{gray}
    HRI  \cite{DBLP:conf/eccv/MaoLS20}& 7.4 & 56.5 & 73.9 & 106.5 & 13.7 & 78.1 & 101.9 & 138.8 & 8.6 & 49.2 & 67.4 & 105.0 & 10.2 & 75.8 & 107.6 & 178.2\\ 
    STDGCN \cite{DBLP:conf/cvpr/MaNLZL22} & 7.5 & 56.0 & 73.3 & 105.9 & 14.0 & 77.3 & 100.2 & $\mathbf{136.4}$ & 8.7 & 48.8 & 66.5 & $\mathbf{102.7}$ & 10.2 & $\underline{73.3}$ & $\underline{102.8}$ & $\underline{167.0}$\\
    \rowcolor{gray}
    MMA  \cite{DBLP:journals/ijcv/MaoLSL21}& 7.2 & $\underline{55.0}$ & $\underline{72.4}$ & $\underline{105.7}$ & 13.6 & $\underline{77.2}$ & 100.5 & 136.7 & 8.4 & $\underline{48.4}$ & 66.5 & 104.6 & 9.8 & 74.9 & 105.8 & 172.9\\
    siMLPe \cite{DBLP:conf/wacv/GuoDSLAM23} & $\underline{6.5}$ & 55.8 & 73.1 & 106.7 & $\underline{12.4}$ & 77.3 & $\underline{99.8}$ & 137.5 & $\underline{8.1}$ & 48.6 & $\underline{66.3}$ & 103.3 & $\underline{8.8}$ & 73.8 & 103.4 & 168.7\\
    \rowcolor{gray}
    \hline
    Ours & $\mathbf{6.3}$ & $\mathbf{52.8}$ & $\mathbf{69.8}$ & $\mathbf{104.4}$ & $\mathbf{11.6}$ & $\mathbf{72.8}$ & $\mathbf{97.7}$ & $\underline{136.5}$ & $\mathbf{7.6}$ & $\mathbf{46.5}$ & $\mathbf{64.6}$ & $\underline{103.3}$ & $\mathbf{8.3}$ & $\mathbf{70.7}$ & $\mathbf{101.0}$ & $\mathbf{166.4}$\\ 
    \hline
    scenarios  & \multicolumn{4}{c|}{purchases} & \multicolumn{4}{c|}{sitting} & \multicolumn{4}{c|}{sitting down} & \multicolumn{4}{c|}{taking photo} \\
    \hline
    milliseconds & 80ms & 400ms & 560ms & 1000ms & 80ms & 400ms & 560ms & 1000ms & 80ms & 400ms & 560ms & 1000ms & 80ms & 400ms & 560ms & 1000ms  \\ 
    \hline
    \rowcolor{gray}
    ResSup \cite{DBLP:conf/cvpr/MartinezB017}& 28.7 & 100.7 & 122.1 & 154.0 & 23.8 & 91.2 & 113.7 & 152.6 & 31.7 & 112.0 & 138.8 & 187.4 & 21.9 & 87.6 & 110.6 & 153.9 \\
    convSeq2Seq \cite{DBLP:conf/cvpr/LiZLL18}& 20.3 & 89.9 & 111.3 & 151.5 & 13.5 & 63.1 & 82.4 & 120.7 & 20.7 & 82.7 & 106.5 & 150.3 & 12.7 & 63.6 & 84.4 & 128.1 \\ 
    \rowcolor{gray}
    LTD-50-25 \cite{DBLP:conf/iccv/MaoLSL19} & 15.2 & 78.1 & 99.2 & 134.9 & 10.4 & 58.3 & 79.2 & 118.7 & 17.1 & 76.4 & 100.2 & 143.8 & 9.6 & 54.3 & 75.3 & 118.8\\
    LTD-10-10 \cite{DBLP:conf/iccv/MaoLSL19} & 13.9 & 75.9 & 99.4 & 135.9 & 9.8 & 55.9 & 78.5 &  118.8 & 15.6 & 71.7 & 96.2 & 142.2 & 8.9 & 51.7 & 72.5 & 116.3\\
    \rowcolor{gray}
    HRI \cite{DBLP:conf/eccv/MaoLS20} & 13.0 & 73.9 & 95.6 & 134.2 & 9.3 & 56.0 & 76.4 & 115.9 & 14.9 & 72.0 & 97.0 & 143.6 & 8.3 & 51.5 & 72.1 & 115.9\\ 
    STDGCN \cite{DBLP:conf/cvpr/MaNLZL22} & 13.2 & 74.0 & 95.7 & $\mathbf{132.1}$ & 9.1 & $\underline{54.6}$ & $\underline{75.1}$ & 114.8 & 14.7 & $\mathbf{70.0}$ & $\mathbf{94.4}$ & $\mathbf{139.0}$ & 8.2 & $\underline{50.2}$ & $\underline{70.5}$ & 112.9\\
    \rowcolor{gray} 
    MMA  \cite{DBLP:journals/ijcv/MaoLSL21} & 12.8 & 72.8 & 94.5 & 133.1 & 9.1 & 55.4 & 75.8 & 115.0 & 14.7 & 71.3 & 96.0 & 141.8 & 8.2 & 51.1 & 71.8 & 115.2\\
    siMLPe \cite{DBLP:conf/wacv/GuoDSLAM23} & $\underline{11.7}$ & $\underline{72.4}$ & $\underline{93.8}$ & $\underline{132.5}$ & $\underline{8.6}$ & 55.2 & 75.4 & $\mathbf{114.1}$ & $\underline{13.6}$ & 70.8 & 95.7 & 142.4 & $\underline{7.8}$ & 50.8 & 71.0 & $\underline{112.8}$\\
    \rowcolor{gray}
    \hline
    Ours & $\mathbf{11.1}$ & $\mathbf{71.7}$ & $\mathbf{93.0}$ & 132.7 & $\mathbf{8.4}$ & $\mathbf{53.8}$ & $\mathbf{74.7}$ & $\underline{114.5}$ & $\mathbf{13.5}$ & $\underline{70.4}$ & $\underline{95.1}$ & $\underline{141.6}$ & $\mathbf{7.5}$ & $\mathbf{49.1}$ & $\mathbf{69.5}$ & $\mathbf{111.6}$\\ 
    \hline
    scenarios  & \multicolumn{4}{c|}{waiting} & \multicolumn{4}{c|}{walking dog} & \multicolumn{4}{c|}{walking together} & \multicolumn{4}{c|}{average} \\
    \hline
    milliseconds & 80ms & 400ms & 560ms & 1000ms & 80ms & 400ms & 560ms & 1000ms & 80ms & 400ms & 560ms & 1000ms & 80ms & 400ms & 560ms & 1000ms  \\ 
    \hline
    \rowcolor{gray}
    ResSup \cite{DBLP:conf/cvpr/MartinezB017}& 23.8 & 87.7 & 105.4 & 135.4 & 36.4 & 110.6 & 128.7 & 164.5 & 20.4 & 67.3 & 80.2 & 98.2 & 25.0 & 88.3 & 106.3 & 136.6  \\ 
    convSeq2Seq \cite{DBLP:conf/cvpr/LiZLL18}& 14.6 & 68.7 & 87.3 & 117.7 & 27.7 & 103.3 & 122.4 & 162.4 & 15.3 & 61.2 & 72.0 & 87.4 & 16.6 & 72.7 & 90.7 & 124.2\\ 
    \rowcolor{gray}
    LTD-50-25 \cite{DBLP:conf/iccv/MaoLSL19} & 10.4 & 59.2 & 77.2 & 108.3 & 22.8 & 88.7 & 107.8 & 156.4 & 10.3 & 46.3 & 56.0 & 65.7 & 12.2 & 61.5 & 79.6 & 112.4\\
    LTD-10-10 \cite{DBLP:conf/iccv/MaoLSL19} & 9.2 & 54.4 & 73.4 & 107.5 & 20.9 & 86.6 & 109.7 & 150.1 & 9.6 & 44.0 & 55.7 & 69.8 & 11.2 & 58.9 & 78.3 & 114.0\\
    \rowcolor{gray}
    HRI \cite{DBLP:conf/eccv/MaoLS20} & 8.7 & 54.9 & 74.5 & 108.2 & 20.1 & 86.3 & 108.2 & 146.9 & 8.9 & 41.9 & 52.7 & 64.9 & 10.4 & 58.3 & 77.3 & 112.1 \\
    STDGCN \cite{DBLP:conf/cvpr/MaNLZL22} & 8.7 & 53.6 & 71.6 & $\underline{103.7}$ & 20.4 & 84.6 & 105.7 & 145.9 & 8.9 & 43.8 & 54.4 & 64.6 & 10.6 & 57.9 & 76.3 & 109.7\\
    \rowcolor{gray}
    MMA  \cite{DBLP:journals/ijcv/MaoLSL21} & 8.4 & 53.8 & 72.7 & 105.1 & 19.6 & $\underline{84.1}$ & $\mathbf{105.1}$ & $\underline{141.4}$ & 8.5 & $\underline{41.1}$ & 51.2 & 63.2 & 10.2 & 57.3 & 75.9 & 110.1\\
    siMLPe \cite{DBLP:conf/wacv/GuoDSLAM23}& $\underline{7.8}$ & $\underline{53.2}$ & $\underline{71.6}$ & 104.6 & $\underline{18.2}$ & $\mathbf{83.6}$ & $\underline{105.6}$ & $\mathbf{141.2}$ & $\underline{8.4}$ & 41.2 & $\underline{50.8}$ & $\mathbf{61.5}$ & $\underline{9.6}$ & $\underline{57.3}$ & $\underline{75.7}$ & $\underline{109.4}$ \\ 
    \rowcolor{gray}
    \hline
    Ours & $\mathbf{7.2}$ & $\mathbf{50.5}$ & $\mathbf{69.0}$ & $\mathbf{103.4}$ & $\mathbf{17.8}$ & 87.0 & 112.0 & 147.5 & $\mathbf{7.7}$ & $\mathbf{38.5}$ & $\mathbf{48.5}$ & $\underline{61.9}$ & $\mathbf{9.1}$ & $\mathbf{55.5}$ & $\mathbf{74.5}$ & $\mathbf{109.2}$ \\ 
    \hline
\end{tabular}}
\end{center}
\caption{Action-wise prediction results on Human3.6M dataset. MPJPE in millimeter at particular frames are reported. 256 samples are tested for each action. The best results are highlighted in \textbf{bold}, and the second best results are \underline{underlined}.} 
\label{tab:h36_both}
\end{table*}
\subsection{Datasets}
\label{subsec::dataset}

\noindent\textbf{Human3.6M} \cite{DBLP:journals/pami/IonescuPOS14} is the most widely used benchmark dataset for human motion prediction.
It contains 15 distinct actions performed by 7 subjects, and each human pose contains 32 joints in exponential map format.
Following \cite{DBLP:conf/iccv/MaoLSL19}, 3D coordinates of the joints are computed based on forward kinematics of the skeleton, and 10 redundant joints are removed. 
For a fair comparison with \cite{DBLP:conf/eccv/MaoLS20, DBLP:conf/iccv/MaoLSL19}, we use subject 11 and subject 5 for validation and testing, and the rest for training.

\noindent\textbf{AMASS} \cite{DBLP:conf/iccv/MahmoodGTPB19} unifies multiple Mocap datasets, such as the popular CMU-MoCap dataset using a shared SMPL \cite{DBLP:journals/tog/LoperM0PB15} parameterization.
As in the Human3.6M case, 3D coordinates are obtained and 18 joints are used following \cite{DBLP:conf/eccv/MaoLS20}.
Other settings including frame-rate, train/validate/test partition and testing protocol are all adjusted to be the same as previous works \cite{DBLP:conf/eccv/MaoLS20, DBLP:journals/ijcv/MaoLSL21} for fair comparisons.

\noindent\textbf{3DPW} \cite{DBLP:conf/eccv/MarcardHBRP18} consists of actions performed in challenging scenes.
Following \cite{DBLP:conf/eccv/MaoLS20}, we only evaluate our model trained on the AMASS dataset on the test partition of 3DPW, aiming to examine the generalization of our method.
All settings are adjusted to be the same as \cite{DBLP:conf/eccv/MaoLS20}.
\begin{table*}[t]\scriptsize
\begin{center}
\resizebox{\linewidth}{!}{
    \begin{tabular}{  |c | cccc|cccc | cccc|cccc | } 
      \hline
      dataset  & \multicolumn{8}{c|}{AMASS} & \multicolumn{8}{c|}{3DPW}  \\
      \hline
      milliseconds & 80ms & 160ms & 320ms & 400ms & 560ms & 720ms & 880ms & 1000ms & 80ms & 160ms & 320ms & 400ms & 560ms & 720ms & 880ms & 1000ms  \\ 
      \hline
      \rowcolor{gray}
      convSeq2Seq \cite{DBLP:conf/cvpr/LiZLL18}& 20.6 &36.9 &59.7 &67.6& 79.0& 87.0 &91.5 &93.5 &18.8 &32.9& 52.0 &58.8 &69.4& 77.0 &83.6 &87.8 \\ 
      LTD-10-10 \cite{DBLP:conf/iccv/MaoLSL19} & $\underline{10.3}$& $\underline{19.3}$ &36.6& 44.6 &61.5& 75.9 &86.2& 91.2& $\underline{12.0}$ & $\underline{22.0}$ &38.9 &46.2& 59.1& 69.1 &76.5& 81.1\\ 
      \rowcolor{gray}
      LTD-10-25 \cite{DBLP:conf/iccv/MaoLSL19} & 11.0 &20.7 &37.8 &45.3 &57.2 &65.7& 71.3& 75.2 &12.6 &23.2& 39.7 &46.6 &57.9& 65.8 &71.5& 75.5\\
      HRI \cite{DBLP:conf/eccv/MaoLS20} &  11.3 &20.7 &35.7& 42.0& 51.7 &58.6 &63.4 &67.2& 12.6 &23.1 &39.0& 45.4 &56.0 &63.6& 69.7 &73.7\\ 
      \rowcolor{gray}
      MMA \cite{DBLP:journals/ijcv/MaoLSL21} & 11.0 & 20.3 & 35.0 & 41.2 & 50.7 & 57.4 & $\underline{61.9}$ & 65.8 & 12.4 & 22.6 & 38.1 & $\underline{44.4}$ & $\underline{54.7}$ & $\underline{62.1}$ & $\underline{67.9}$ & $\underline{71.8}$ \\
    siMLPe \cite{DBLP:conf/wacv/GuoDSLAM23} &  10.8& 19.6 &$\underline{34.3}$& $\underline{40.5}$ &$\underline{50.5}$& $\underline{57.3}$& 62.4 &$\underline{65.7}$& 12.1 &22.1 &$\underline{38.1}$& 44.5& 54.9 &62.4& 68.2 &72.2\\ 
    \rowcolor{gray}
      Ours  &  $\mathbf{9.8}$ &	$\mathbf{17.8}$&	$\mathbf{31.8}$&	$\mathbf{37.9}$&	$\mathbf{47.8}$&	$\mathbf{55.3}$&	$\mathbf{60.7}$&	$\mathbf{65.4}$& $\mathbf{11.6}$	&$\mathbf{20.8}$&	$\mathbf{36.4}$&	$\mathbf{42.8}$	&$\mathbf{53.3}$&	$\mathbf{61.0}$&	$\mathbf{66.9}$&	$\mathbf{71.0}$\\ 
      \hline
    \end{tabular}}
\end{center}
\caption{Short-term and long-term prediction results on AMASS and 3DPW datasets. The models are trained on the AMASS dataset. MPJPE in millimeter at particular frames are reported. The best results are highlighted in \textbf{bold}, and the second best results are \underline{underlined}.} 
\label{tab:amass_3dpw}
\end{table*}
\subsection{Experimental Setup}
\label{subsec::setup}

\noindent\textbf{Evaluation metrics.} We use the \textit{Mean Per Joint Position Error} (MPJPE) for performance evaluation, which is the most widely used evaluation metric for 3D human motion prediction.
Specifically, MPJPE calculates the averaged $\ell_2$ difference between the predicted pose sequence and the corresponding groundtruth across all joints.
As in previous works \cite{DBLP:conf/eccv/MaoLS20, DBLP:conf/iccv/MaoLSL19}, we report MPJPE at individual frames.

\noindent\textbf{Baselines.} We compare our model performance with one RNN-based method, ResSup \cite{DBLP:conf/cvpr/MartinezB017}, one CNN-based method convSeq2Seq \cite{DBLP:conf/cvpr/LiZLL18}, and multiple feed-forward methods, including LTD \cite{DBLP:conf/iccv/MaoLSL19}, HRI \cite{DBLP:conf/eccv/MaoLS20}, STDGCN \cite{DBLP:conf/cvpr/MaNLZL22}, MMA \cite{DBLP:journals/ijcv/MaoLSL21} and siMLPe \cite{DBLP:conf/wacv/GuoDSLAM23}, constituting the state of the art.
The results of MMA are taken from \cite{DBLP:journals/ijcv/MaoLSL21}, and all other results are taken from \cite{DBLP:conf/wacv/GuoDSLAM23}.
Note that LTD results are based on different configurations: the numbers indicate observed frames and predicted frames respectively during training.

\noindent\textbf{Implementation details.} We use PyTorch \cite{DBLP:journals/corr/abs-1912-01703} to implement~\name.
We set input length $H=50$, query length $L=10$ and output length $F=10$ for Human3.6M and $F=25$ for AMASS and 3DPW datasets following prior works \cite{DBLP:conf/eccv/MaoLS20, DBLP:conf/wacv/GuoDSLAM23} for fair comparisons.

For the motion attention module, $f_q(\cdot)$ and $f_k(\cdot)$ are implemented as 2-layer CNNs. 
For the motion refinement module, we iteratively perform $N=3$ learning stages, each of which contains 
$K=5$ GLBs ($M=2$) with dropout ratio 0.3.
We set $d=256$ as latent representation dimension across different components of \name.

Adam~\cite{DBLP:journals/corr/KingmaB14}~is used as the optimizer for all experiments, with $0.005$ as the initial learning rate.
For Human3.6M dataset, the model is trained for $200$ epochs with $32$ as the batch size, and the learning rate is multiplied by $0.97$ at each epoch.
For AMASS and 3DPW datasets, we train the model for $250$ epochs with batch size $128$, and the learning rate is multiplied by $0.98$ at each epoch.
We use NVIDIA V100 GPUs for all experiments.
For more model implementation details, please refer to the supplementary material.
\begin{figure}[!tb]
\centering
\includegraphics[width=0.77\linewidth]{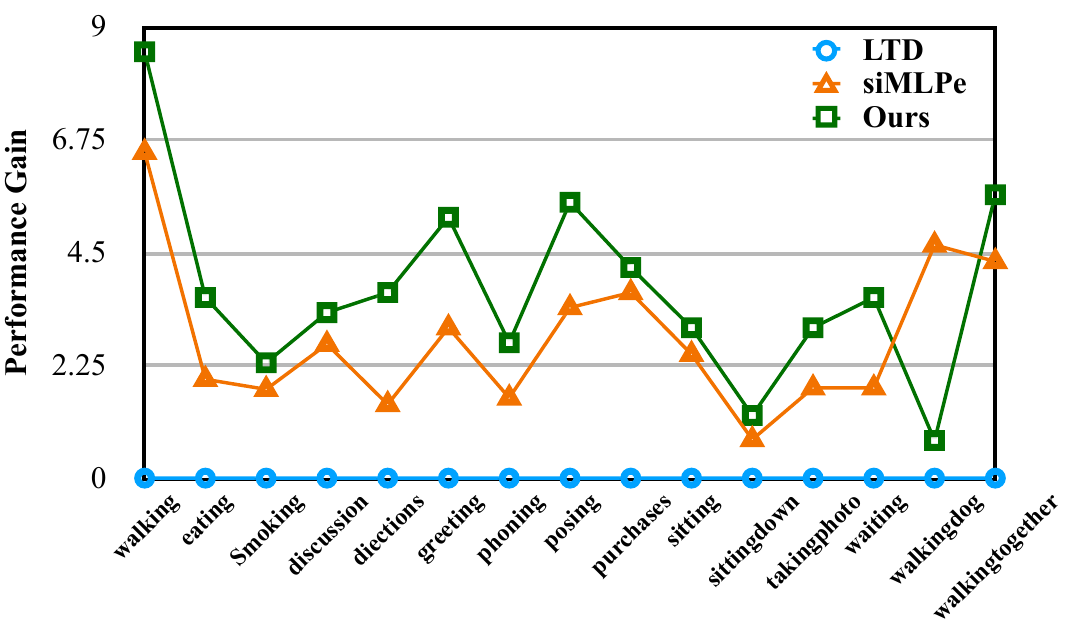}
\caption{Action-wise performance gains for Human3.6M.}
\label{fig:act_wise}
\end{figure}
\subsection{Results}
\label{subsec::result}

In this section, we compare \name~with the selected baselines on various datasets.
We report the MPJPE evaluated at specific frames in millimeters as in \cite{DBLP:conf/eccv/MaoLS20, DBLP:conf/wacv/GuoDSLAM23}.

\noindent\textbf{Human3.6M.} For Human3.6M, we quantitatively report the performance comparison for short-term prediction and long-term prediction in \cref{tab:h36_both}.
Tables with more details are provided in the supplementary material.
\name~outperforms selected baselines on average for both short-term and long-term prediction by significant margins.
In \cref{fig:act_wise}, we also show the action-wise performance gain for \name~and siMLPe with respect to LTD.
We can observe that our method demonstrates large performance gains for actions with repetitive patterns such as ``eating", taking advantages of the motion summary generated by the motion attention module.
However, we interestingly note that our method does not perform as well on the "walkingdog" action as other actions.
We believe this is due to the inherently large periodicity of this particular action, resulting in the current motion summary not helping the model to make better predictions.
We validate our hypothesis in the last part of the ablation study.

\begin{figure}[!tb]
\centering
\includegraphics[width=0.9\linewidth]{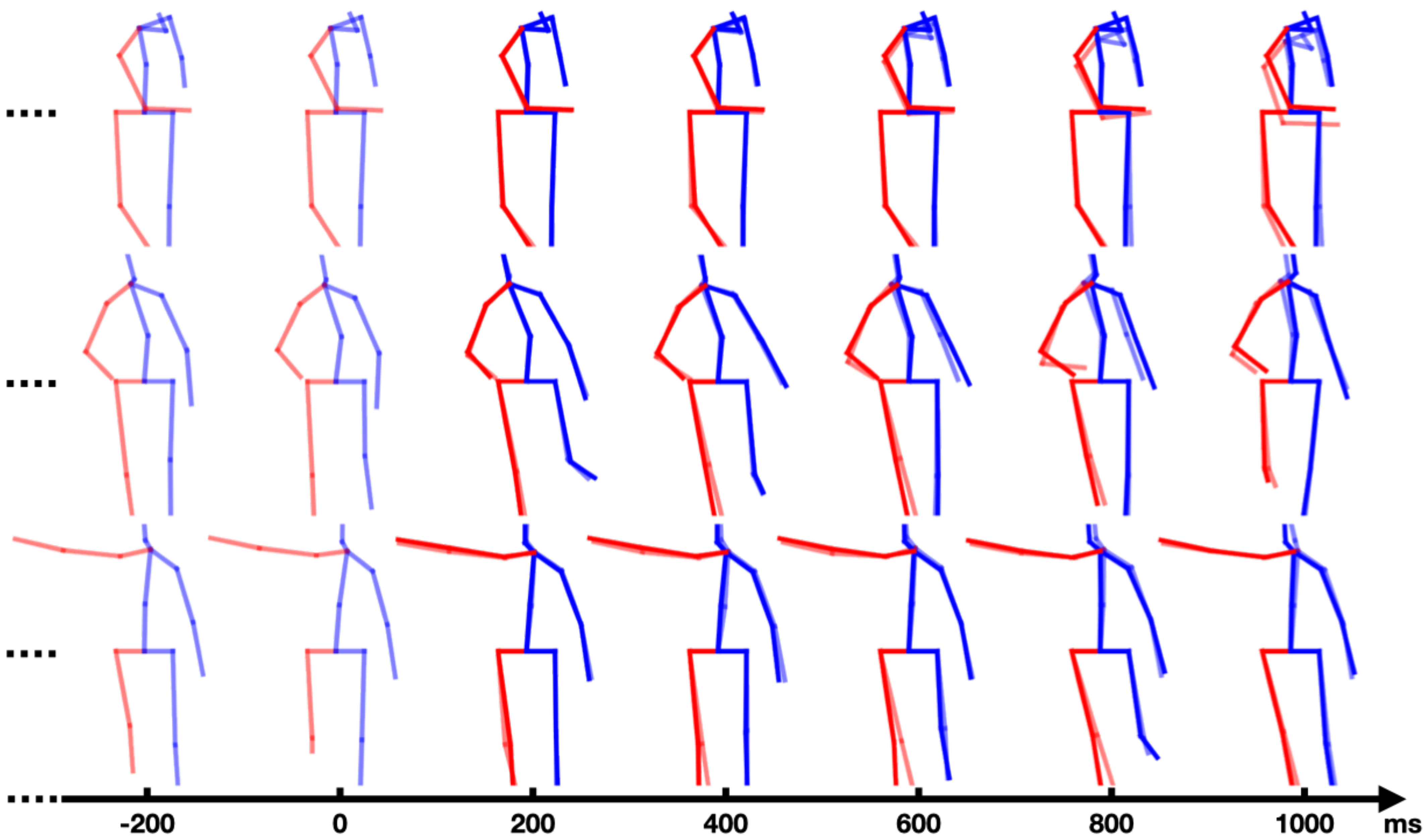}
\caption{Qualitative comparisons with groundtruth for ``sitting" (top), ``walking" (middle) and ``walkingtogether" (bottom) actions from Human3.6M. Solid lines represent predictions, and transparent lines represent groundtruth.}
\label{fig:quality}
\end{figure}

In terms of qualitative evaluation, in \cref{fig:quality}, we show the visualized predicted motion with respect to the correspond groundtruth motion for action ``eating", ``walking" and ``walkingdog" of Human3.6M.
We can see that the predictions generated from \name~match the groundtruth motion consistently in both shot-term and long-term. 

\noindent\textbf{AMASS and 3DPW.} For AMASS and 3DPW, we report performance comparisons quantitatively at particular frames in \cref{tab:amass_3dpw}.
As previously mentioned, the model is trained on the AMASS dataset and tested on the test partition of AMASS and 3DPW separately as in \cite{DBLP:conf/eccv/MaoLS20, DBLP:conf/wacv/GuoDSLAM23}.
In both experiments, \name~consistently outperforms selected baselines across all evaluated frames by large margins.
The superior performance on 3DPW further demonstrates the robustness of our proposed method.

\subsection{Ablation Analysis}
\label{subsec::ablation}

In this section, We conduct an ablation study on Human3.6M dataset to investigate how different components of~\name~may affect its motion modeling ability.
%
%
%
%
\begin{table}[t]\scriptsize
\begin{center}
\resizebox{\linewidth}{!}{
    \begin{tabular}{  |c | cccc | c | } 
      \hline
      variant & 80ms & 400ms & 560ms & 1000ms & average \\
      \hline
      \rowcolor{gray}
      single stage, no attention, no loss guidance & 21.3& 79.9& 104.6& 143.1 & 87.2\\
      no motion attention guidance & 18.9& 78.6& 101.8& 139.1 & 84.6\\
      \rowcolor{gray}
      without iterative refinement & 9.8 & 56.4 & 75.3 & 110.1 & 62.9 \\
      without multi-component loss & 9.7 & 56.4 & 75.5 & 109.8 & 62.9 \\
      \hline 
      \rowcolor{gray}
      full model & $\textbf{9.1}$ & $\mathbf{55.5}$ &$\mathbf{74.5}$&	$\mathbf{109.2}$ & $\mathbf{62.1}$ \\
      \hline
    \end{tabular}}
\end{center}
\caption{Ablation on \name's architecture.} 
\label{tab:mdl_all}
\end{table}
\begin{table}[t]\scriptsize
\begin{center}
\resizebox{\linewidth}{!}{
    \begin{tabular}{ | c | cccccccc | } 
    \hline
      N-M & 80ms & 160ms & 320ms & 400ms & 560ms & 720ms & 880ms & 1000ms \\ 
      \hline
    \rowcolor{gray}
     1-6 & 9.8&	21.7&	45.5&	56.4&	75.3&	90.0&	102.2&	110.1 \\ 
     2-3 & 9.3&	21.1&	45.0&	55.8&	74.8&	89.5&	101.6&	109.5 \\ 
     \rowcolor{gray}
     6-1 & 9.1&	20.7&	44.6&	55.7&	74.9&	89.6&	101.7&	109.5 \\ 
     \hline
     4-2 & 9.1 & 20.8 & 44.9 & 56.0 & 75.1 & 89.8 & 101.9 & 109.9 \\
     \rowcolor{gray}
     5-2 & 9.0 & 20.7 & 44.8 & 55.8 & 74.9 & 89.7 & 101.8 & 109.7 \\
     6-2 & $\textbf{8.9}$ & 20.7 & 44.9 & 55.9 & 74.9 & 89.5 & 101.5 & 109.4 \\
    \hline
     \rowcolor{gray}
      3-2 (ours) & 9.1 & $\mathbf{20.7}$ & $\mathbf{44.5}$ & $\mathbf{55.5}$ &$\mathbf{74.5}$&	$\mathbf{89.2}$	&$\mathbf{101.3}$&	$\mathbf{109.2}$ \\
     \hline
    \end{tabular}}
\end{center}
\caption{Ablation on learning stage configurations.} 
\label{tab:abla_sm_lr}
\end{table}
\begin{figure}[!tb]
\centering
\includegraphics[width=0.76\linewidth]{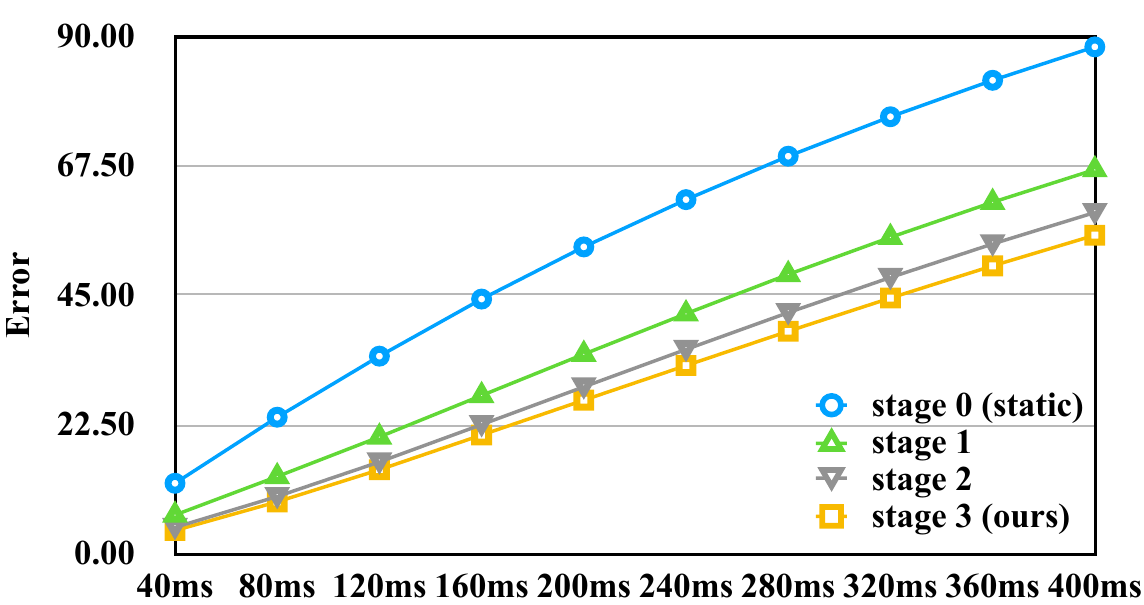}
\caption{Frame-wise error across different refinement stages: 0 (static) to 4 (ours) for Human3.6M.}
\label{fig:stage_improve}
\end{figure}
%
\noindent\textbf{Model architecture.} We first show how each design choice improves the effectiveness of \name, including 1) the motion attention module 2) the multi-stage motion refinement module, and 3) the multi-component loss.
All ablation results are shown in \cref{tab:mdl_all}.
Recall that the full model has $N=3$ refining stages each containing $M=2$ GLM stages, denoting as 3-2 configuration.
The averaged prediction error is 62.1.
We first establish a plain baseline model which does not utilize the proposed attention module, refinement module and multi-component loss.
The prediction error drastically rises to 87.23.
1) To validate the effectiveness of motion attention guidance, we replace the motion attention module with the ``Copy" operator \cite{DBLP:conf/cvpr/MaNLZL22} in the full model, leading to a large prediction error of 84.61.
2) Instead of using 3-2 configuration, we use 1-6 configuration, corresponding to a single-stage model with similar computational power.
The prediction error becomes 62.9.
3) Lastly, we use the simple $\ell_2$ loss to supervise \name, also yielding a 62.9 prediction error.

%


\noindent\textbf{Number of learning stages.} In \cref{tab:abla_sm_lr}, we study the specific configuration of the motion refinement module.
We first examine configurations with similar computational power, and then gradually increase the number of refinements performed while keeping the number of GLMs used per refinement stage fixed.
As shown in the table, \name~achieves the best performance with the 3-2 configuration.

\noindent\textbf{Loss components.} In \cref{tab:abla_aug_loss}, we conduct a detailed ablation study on the proposed loss components.
From the table, we observe that all components help \name~achieve better performance across all frames, while each component has its own focus.
In particular, the spatial-temporal-aware loss puts more emphasis on short-term prediction, while the velocity-aware loss focuses more on long-term improvement.
Besides, the global-smoothness-aware loss uniformly improves the global performance.
It is evident that the proposed loss is indispensable to \name, especially the spatial-temporal-aware loss that considers human kinematic structure. 


\noindent\textbf{Benefit of refinement.} To further validate and  demonstrate the effectiveness of the motion refinement module, we show the frame-wise error of motion generated from each stage of refinement, \ie, stage-wise model outputs from $N=0$ to $N=3$ with respect to the groundtruth in \cref{fig:stage_improve}.
We can observe that the error decreases steadily at each stage.
This indicates that the motion refinement module successfully predicts better motion as it iteratively generates refined motion that is closer to the groundtruth, greatly reducing the learning difficulty.

\noindent\textbf{Arbitrary length history handling.}
To demonstrate \name's ability of handling inputs of arbitrary length, we show the results of feeding different numbers of observed frames of into to \name~during testing in \cref{tab:abla_lh}.
From the table, we can see that the benefits of using longer history motion becomes more evident in long prediction horizon.

%% file: files/conclusion.tex
\section{Conclusion}

In this work, we present~\name, a novel framework for human motion prediction.
Specifically, we propose 1) a motion attention module that generates motion summary, and 2) a motion refinement module that predicts future motion in a multi-stage manner using the latest observed motion and motion summary.
We also propose 3) a multi-component loss which is spatial-temporal-aware, velocity-aware and global-smoothness-aware, guiding the model to predict realistic motion in both short-term and long-term.
The results obtained from extensive experiments and detailed analysis show that~\name~has significant performance gains over state-of-the-art baselines on benchmark datasets, demonstrating the effectiveness of our method.
\begin{table}[t]\scriptsize
\begin{center}
\resizebox{\linewidth}{!}{
    \begin{tabular}{ | c | cccccccc | } 
    \hline
      variant & 80ms & 160ms & 320ms & 400ms & 560ms & 720ms & 880ms & 1000ms \\ 
      \hline
     w/o s.t. weights & 9.4	&21.2&	45.3&	56.2&	75.1&	89.7&	101.7&	109.4\\
     \rowcolor{gray}
     w/o $\mathcal{L}_v$ & 9.1&	20.7&	44.6&	55.7&	75.1&	89.8&	101.9&	109.7\\
     w/o query reconst. & 9.2&	20.9&	44.9&	55.8&	74.8&	89.5&	101.8&	109.7\\
     \hline
     \rowcolor{gray}
    $\mathcal{L}_{st} + \mathcal{L}_v$ (ours) & $\mathbf{9.1}$ & $\mathbf{20.7}$ & $\mathbf{44.5}$ & $\mathbf{55.5}$ &$\mathbf{74.5}$&	$\mathbf{89.2}$	&$\mathbf{101.3}$&	$\mathbf{109.2}$	 \\ 
     \hline
    \end{tabular}}
\end{center}
\caption{Ablation on loss configurations.} 
\label{tab:abla_aug_loss}
\end{table}
\begin{table}[t]\scriptsize
\begin{center}
\resizebox{\linewidth}{!}{
    \begin{tabular}{ | c | cccccccc | } 
    \hline
      variant & 80ms & 160ms & 320ms & 400ms & 560ms & 720ms & 880ms & 1000ms \\ 
      \hline
     ours-50 & 17.8	& $\mathbf{37.9}$	& $\mathbf{72.8}$ & $\mathbf{87.0}$ & 112.0 & 124.2 & 137.7 & 147.5 \\
     \hline
     \rowcolor{gray}
     ours-100 & $\mathbf{17.7}$ & 38.0 & 73.5 & 87.4 & $\mathbf{111.5}$ & $\mathbf{121.9}$ & $\mathbf{134.8}$ & $\mathbf{145.1}$ \\
     \hline
    \end{tabular}}
\end{center}
\caption{Results on feeding selected sequences: ``walkingdog" action that contain similar patterns in longer history. Number indicates observed frames during testing.} 
\label{tab:abla_lh}
\end{table}

%% file: files/appendix.tex
\clearpage
\appendix

\section{Model Implementation Details}

We use PyTorch to implement~\name.

For the motion attention module, $f_q(\cdot)$ and $f_k(\cdot)$ are implemented as 2-layer CNNs. 
Specifically, the kernel size of these two layers are set to 6 and 5 respectively with intermediate ReLU \cite{DBLP:conf/icml/NairH10} activation.
These kernel size choices without padding provide us with a receptive field of 10 which corresponds to the query length $L$. 

For the motion refinement module, we iteratively perform $N=3$ learning stages, each of which contains 
$K=5$ GLBs ($M=2$).
Each GLB sequentially executes GC, batch normalization \cite{DBLP:conf/icml/IoffeS15}, tanh activation and dropout \cite{DBLP:journals/jmlr/SrivastavaHKSS14}.
The dropout ratio is set to 0.3.
We set $d=256$ as the latent representation dimension for both motion attention module and motion refinement module.

Adam~\cite{DBLP:journals/corr/KingmaB14}~is used as the optimizer for all experiments, with $0.005$ as the initial learning rate.
For Human3.6M dataset, the model is trained for $200$ epochs with $32$ as the batch size, and the learning rate is multiplied by $0.97$ at each epoch.
For AMASS and 3DPW datasets, we train the model for $250$ epochs with batch size $128$, and the learning rate is multiplied by $0.98$ at each epoch.
\name~has around 2.30 million parameters for training on Human3.6M dataset.
We use NVIDIA V100 GPUs to train \name~for all experiments.

\section{More Results for Human3.6M Dataset}

\begin{table*}[t]\scriptsize
\begin{center}
\resizebox{\linewidth}{!}{
    \begin{tabular}{ |c |cccc| cccc | cccc| cccc| } 
    \hline
    scenarios  & \multicolumn{4}{c|}{walking} & \multicolumn{4}{c|}{eating} & \multicolumn{4}{c|}{smoking} & \multicolumn{4}{c|}{discussion} \\
    \hline
    milliseconds & 80ms & 160ms & 320ms & 400ms & 80ms & 160ms & 320ms & 400ms & 80ms & 160ms & 320ms & 400ms & 80ms & 160ms & 320ms & 400ms  \\ 
    \hline
    \rowcolor{gray}
    Res. Sup.\cite{DBLP:conf/cvpr/MartinezB017} & 23.2  & 40.9 & 61.0 & 66.1 & 16.8 & 31.5 & 53.5 & 61.7 & 18.9 & 34.7 & 57.5 & 65.4 & 25.7 & 47.8 & 80.0 & 91.3   \\ 
    convSeq2Seq \cite{DBLP:conf/cvpr/LiZLL18}& 17.7 &33.5 &56.3 &63.6 &11.0 &22.4 &40.7& 48.4 &11.6 &22.8& 41.3 &48.9 &17.1 &34.5& 64.8& 77.6 \\ 
    \rowcolor{gray}
    LTD  \cite{DBLP:conf/iccv/MaoLSL19}& 11.1 &21.4& 37.3& 42.9 &7.0 &14.8& 29.8& 37.3& 7.5 &15.5 &30.7 &37.5 &10.8 &24.0 &52.7& 65.8 \\ 
    HRI \cite{DBLP:conf/eccv/MaoLS20} & 10.0 &19.5& 34.2 &39.8 &6.4 &14.0& 28.7 &36.2 &7.0 &14.9 &29.9 &36.4 &10.2 &23.4 &52.1& 65.4 \\ 
    \rowcolor{gray}
    MMA  \cite{DBLP:journals/ijcv/MaoLSL21} & 9.9 & 19.3&  33.7 & 39.0 & 6.2 & 13.7 & 28.1 & 35.3 & 6.8&  14.5 & 29.0 & 35.5 & 9.9 & 22.8 & 51.0 & 64.0 \\
    siMLPe \cite{DBLP:conf/wacv/GuoDSLAM23} & 9.9 & - & - &  39.6 & 5.9 & - & - & 36.1 & 6.5 & - & - & 36.3  & 9.4 & - & - & 64.3 \\ 
    \rowcolor{gray}
    Ours & $\mathbf{9.0}$ & $\mathbf{17.9}$ & $\mathbf{31.6}$ & $\mathbf{36.7}$ & $\mathbf{5.6}$ & $\mathbf{12.7}$ & $\mathbf{26.7}$ & $\mathbf{33.9}$ & $\mathbf{6.2}$ & $\mathbf{13.5}$ & $\mathbf{28.1}$ & $\mathbf{35.1}$ & $\mathbf{8.7}$ & $\mathbf{21.4}$ & $\mathbf{49.7}$ & $\mathbf{62.9}$ \\ 
    \hline
    scenarios  & \multicolumn{4}{c|}{directions} & \multicolumn{4}{c|}{greeting} & \multicolumn{4}{c|}{phoning} & \multicolumn{4}{c|}{posing} \\
    \hline
    milliseconds & 80ms & 160ms & 320ms & 400ms & 80ms & 160ms & 320ms & 400ms & 80ms & 160ms & 320ms & 400ms & 80ms & 160ms & 320ms & 400ms  \\ 
    \hline
    \rowcolor{gray}
    Res. Sup. \cite{DBLP:conf/cvpr/MartinezB017}& 21.6 & 41.3 &72.1 &84.1 &31.2& 58.4& 96.3& 108.8& 21.1& 38.9 &66.0& 76.4& 29.3& 56.1& 98.3& 114.3   \\ 
    convSeq2Seq \cite{DBLP:conf/cvpr/LiZLL18}& 13.5 &29.0& 57.6& 69.7& 22.0& 45.0& 82.0& 96.0& 13.5& 26.6& 49.9& 59.9& 16.9& 36.7& 75.7& 92.9 \\
     \rowcolor{gray}
    LTD  \cite{DBLP:conf/iccv/MaoLSL19}&  8.0 &18.8& 43.7& 54.9& 14.8& 31.4& 65.3& 79.7& 9.3 &19.1& 39.8& 49.7& 10.9& 25.1& 59.1& 75.9 \\ 
    HRI \cite{DBLP:conf/eccv/MaoLS20} & 7.4 &18.4 &44.5 &56.5& 13.7& 30.1& 63.8& 78.1& 8.6 &18.3& 39.0& 49.2& 10.2& 24.2& 58.5& 75.8 \\ 
    \rowcolor{gray}
    MMA  \cite{DBLP:journals/ijcv/MaoLSL21} & 7.2 & 18.0 & 43.4 & 55.0&  13.6&  29.9 & 62.9 & 77.2 & 8.4 & 18.0&  38.3&  48.4 & 9.8 & 23.7 & 57.8 & 74.9 \\
    siMLPe \cite{DBLP:conf/wacv/GuoDSLAM23} & 6.5 &- &- & 55.8 & 12.4 &- &- & 77.3 & 8.1& -&-& 48.6 &8.8&-&-& 73.8\\ 
    \rowcolor{gray}
    Ours & $\mathbf{6.3}$ & $\mathbf{16.8}$ & $\mathbf{41.6}$ & $\mathbf{52.8}$ & $\mathbf{11.6}$ & $\mathbf{27.0}$ & $\mathbf{58.9}$ & $\mathbf{72.8}$ & $\mathbf{7.6}$ & $\mathbf{16.9}$ & $\mathbf{36.8}$ & $\mathbf{46.5}$ & $\mathbf{8.3}$ & $\mathbf{21.6}$ & $\mathbf{54.3}$ & $\mathbf{70.7}$ \\ 
    \hline
    scenarios  & \multicolumn{4}{c|}{purchases} & \multicolumn{4}{c|}{sitting} & \multicolumn{4}{c|}{sitting down} & \multicolumn{4}{c|}{taking photo} \\
    \hline
    milliseconds & 80ms & 160ms & 320ms & 400ms & 80ms & 160ms & 320ms & 400ms & 80ms & 160ms & 320ms & 400ms & 80ms & 160ms & 320ms & 400ms  \\ 
    \hline
    \rowcolor{gray}
    Res. Sup. \cite{DBLP:conf/cvpr/MartinezB017}& 28.7 &52.4 &86.9& 100.7& 23.8& 44.7& 78.0& 91.2& 31.7& 58.3& 96.7& 112.0& 21.9& 41.4& 74.0& 87.6  \\
    convSeq2Seq \cite{DBLP:conf/cvpr/LiZLL18}& 20.3 &41.8 &76.5 &89.9 &13.5 &27.0 &52.0& 63.1 &20.7 &40.6& 70.4 &82.7& 12.7& 26.0& 52.1 &63.6 \\ 
    \rowcolor{gray}
    LTD \cite{DBLP:conf/iccv/MaoLSL19}& 13.9 &30.3 &62.2& 75.9& 9.8& 20.5 &44.2& 55.9 &15.6& 31.4 &59.1& 71.7& 8.9& 18.9 &41.0& 51.7\\ 
    HRI \cite{DBLP:conf/eccv/MaoLS20} & 13.0 &29.2& 60.4& 73.9 &9.3& 20.1& 44.3 &56.0& 14.9 &30.7& 59.1& 72.0& 8.3& 18.4& 40.7& 51.5 \\ 
    \rowcolor{gray}
    MMA  \cite{DBLP:journals/ijcv/MaoLSL21} & 12.8 & 28.7 & 59.4 & 72.8 & 9.1&  19.7&  43.7 & 55.4 & 14.7 & 30.4 & 58.4 & 71.3&  8.2 & 18.1 & 40.2 & 51.1 \\
    siMLPe \cite{DBLP:conf/wacv/GuoDSLAM23} & 11.7&-&-& 72.4&8.6&-&-& 55.2&13.6 &-&-&70.8 & 7.8&-&-& 50.8\\ 
    \rowcolor{gray}
    Ours & $\mathbf{11.1}$ & $\mathbf{26.8}$ & $\mathbf{58.2}$ & $\mathbf{71.7}$ & $\mathbf{8.4}$ & $\mathbf{18.8}$ & $\mathbf{42.2}$ & $\mathbf{53.8}$ & $\mathbf{13.5}$ & $\mathbf{29.2}$ & $\mathbf{57.5}$ & $\mathbf{70.4}$ & $\mathbf{7.5}$ & $\mathbf{17.1}$ & $\mathbf{38.6}$ & $\mathbf{49.1}$ \\ 
    \hline
    scenarios  & \multicolumn{4}{c|}{waiting} & \multicolumn{4}{c|}{walking dog} & \multicolumn{4}{c|}{walking together} & \multicolumn{4}{c|}{average} \\
    \hline
    milliseconds & 80ms & 160ms & 320ms & 400ms & 80ms & 160ms & 320ms & 400ms & 80ms & 160ms & 320ms & 400ms & 80ms & 160ms & 320ms & 400ms  \\ 
    \hline
    \rowcolor{gray}
    Res. Sup. \cite{DBLP:conf/cvpr/MartinezB017}& 23.8 &44.2 &75.8& 87.7& 36.4 &64.8 &99.1 &110.6& 20.4 &37.1& 59.4 &67.3 &25.0& 46.2& 77.0& 88.3  \\ 
    convSeq2Seq \cite{DBLP:conf/cvpr/LiZLL18}& 14.6 &29.7 &58.1 &69.7& 27.7& 53.6 &90.7 &103.3 &15.3& 30.4& 53.1& 61.2 &16.6 &33.3& 61.4 &72.7 \\ 
    \rowcolor{gray}
    LTD \cite{DBLP:conf/iccv/MaoLSL19} & 9.2 &19.5 &43.3 &54.4& 20.9& 40.7 &73.6 &86.6& 9.6 &19.4 &36.5 &44.0& 11.2 &23.4 &47.9 &58.9 \\
    HRI \cite{DBLP:conf/eccv/MaoLS20} & 8.7 &19.2& 43.4& 54.9 &20.1 &40.3& 73.3& 86.3& 8.9 &18.4 &35.1 &41.9& 10.4 &22.6& 47.1 &58.3 \\
    \rowcolor{gray}
    MMA  \cite{DBLP:journals/ijcv/MaoLSL21} & 8.4 & 18.7 & 42.5&  53.8 & 19.6&  39.5 & 71.7&  84.1&  8.5 & 17.9 & 34.3 & 41.1 & 10.2&  22.2&  46.3 & 57.3 \\
    siMLPe \cite{DBLP:conf/wacv/GuoDSLAM23} & 7.8&-&-& 53.2&18.2&-&-& $\mathbf{83.6}$ & 8.4&-&-& 41.2 &9.6 &21.7&46.3&57.3 \\ 
    \rowcolor{gray}
    Ours & $\mathbf{7.2}$ & $\mathbf{17.0}$ & $\mathbf{39.6}$ & $\mathbf{50.5}$ & $\mathbf{17.8}$ & $\mathbf{37.9}$ & $\mathbf{72.8}$ & 87.0 & $\mathbf{7.7}$ & $\mathbf{16.2}$ & $\mathbf{31.8}$ & $\mathbf{38.5}$ & $\mathbf{9.1}$ & $\mathbf{20.7}$ & $\mathbf{44.5}$ & $\mathbf{55.5}$ \\ 
    \hline
\end{tabular}}
\end{center}
\caption{Short-term prediction on Human3.6M dataset. MPJPE in millimeter at particular frames (80ms, 160ms, 320ms, 400ms) are reported. 256 samples are tested for each action. The best results are highlighted in \textbf{bold}.} 
\label{tab:h36_short}
\end{table*}
\begin{table*}[t]\footnotesize
\begin{center}
\resizebox{\linewidth}{!}{
    \begin{tabular}{ |c |cccc| cccc | cccc| cccc| } 
    \hline
    scenarios  & \multicolumn{4}{c|}{walking} & \multicolumn{4}{c|}{eating} & \multicolumn{4}{c|}{smoking} & \multicolumn{4}{c|}{discussion} \\
    \hline
    milliseconds & 560ms & 720ms & 880ms & 1000ms & 560ms & 720ms & 880ms & 1000ms & 560ms & 720ms & 880ms & 1000ms & 560ms & 720ms & 880ms & 1000ms  \\ 
    \hline
    \rowcolor{gray}
    Res. Sup.\cite{DBLP:conf/cvpr/MartinezB017} & 71.6&72.5 &76.0& 79.1& 74.9 &85.9& 93.8 &98.0& 78.1& 88.6 &96.6 &102.1 &109.5& 122.0 &128.6 &131.8   \\ 
    convSeq2Seq \cite{DBLP:conf/cvpr/LiZLL18}& 72.2& 77.2& 80.9& 82.3& 61.3& 72.8 &81.8 &87.1 &60.0 &69.4 &77.2& 81.7 &98.1 &112.9& 123.0& 129.3 \\ 
    \rowcolor{gray}
    LTD  \cite{DBLP:conf/iccv/MaoLSL19}& 53.1 &59.9& 66.2 &70.7& 51.1 &62.5& 72.9 &78.6 &49.4 &59.2& 66.9 &71.8& 88.1& 104.5& 115.5 &121.6 \\ 
    HRI \cite{DBLP:conf/eccv/MaoLS20} & 47.4& 52.1 &55.5& 58.1& 50.0& 61.4& 70.6& 75.7 &47.6 &56.6 &64.4 &69.5 &86.6& 102.2& 113.2& 119.8 \\ 
    \rowcolor{gray}
    MMA  \cite{DBLP:journals/ijcv/MaoLSL21} &46.2 & 51.0 & 54.4 & 57.1 & 48.6 & 59.9 & 68.9 & 73.7 & 46.5 & 55.5 & 63.4 & 68.7 & 85.2 & 100.9 & 111.6 & 117.5 \\
    siMLPe \cite{DBLP:conf/wacv/GuoDSLAM23} & 46.8 & - & - &  55.7 & 49.6 & - & - & 74.5 & 47.2 & - & - & 69.3  & 85.7 & - & - & $\mathbf{116.3}$\\ 
    \rowcolor{gray}
    Ours & $\mathbf{44.5}$ & $\mathbf{48.9}$ & $\mathbf{51.8}$ & $\mathbf{54.1}$ & $\mathbf{47.2}$ & $\mathbf{58.6}$ & $\mathbf{67.9}$ & $\mathbf{73.1}$ & $\mathbf{46.9}$ & $\mathbf{56.1}$ & $\mathbf{64.1}$ & $\mathbf{69.2}$ & $\mathbf{84.7}$ & $\mathbf{100.6}$ & $\mathbf{110.9}$ & 117.1 \\ 
    \hline
    scenarios  & \multicolumn{4}{c|}{directions} & \multicolumn{4}{c|}{greeting} & \multicolumn{4}{c|}{phoning} & \multicolumn{4}{c|}{posing} \\
    \hline
    milliseconds & 560ms & 720ms & 880ms & 1000ms & 560ms & 720ms & 880ms & 1000ms & 560ms & 720ms & 880ms & 1000ms & 560ms & 720ms & 880ms & 1000ms  \\ 
    \hline
    \rowcolor{gray}
    Res. Sup. \cite{DBLP:conf/cvpr/MartinezB017}&  101.1& 114.5 &124.5& 129.1& 126.1 &138.8 &150.3 &153.9& 94.0& 107.7& 119.1& 126.4& 140.3 &159.8 &173.2 &183.2   \\ 
    convSeq2Seq \cite{DBLP:conf/cvpr/LiZLL18}&  86.6 &99.8 &109.9 &115.8 &116.9& 130.7 &142.7& 147.3 &77.1& 92.1 &105.5& 114.0& 122.5& 148.8 &171.8 &187.4  \\
    \rowcolor{gray}
    LTD  \cite{DBLP:conf/iccv/MaoLSL19}&  72.2 &86.7& 98.5& 105.8 &103.7& 120.6& 134.7 &140.9 &67.8 &83.0& 96.4& 105.1 &107.6 &136.1& 159.5& 175.0 \\ 
    HRI \cite{DBLP:conf/eccv/MaoLS20} & 73.9& 88.2& 100.1 &106.5 &101.9 &118.4 &132.7 &138.8& 67.4& 82.9 &96.5& 105.0& 107.6 &136.8& 161.4 &178.2  \\ 
    \rowcolor{gray}
    MMA  \cite{DBLP:journals/ijcv/MaoLSL21} &72.4 & 87.4 & 99.3 & 105.7 & 100.5 & 116.5 & 130.7 & 136.7 & 66.5 & 82.3 & 95.8 & 104.6 & 105.8 & 134.1 & 157.5 & 172.9 \\
    siMLPe \cite{DBLP:conf/wacv/GuoDSLAM23} & 73.1 &- &- & 106.7 & 99.8 &- &- & 137.5 & 66.3 & -&-& 103.3 & 103.4 &-&-& 168.7\\ 
    \rowcolor{gray}
    Ours & $\mathbf{69.8}$&	$\mathbf{84.9}$& $\mathbf{97.2}$ & $\mathbf{104.4}$ & $\mathbf{97.7}$ & $\mathbf{115.9}$ & $\mathbf{129.9}$ & $\mathbf{136.5}$ & $\mathbf{64.6}$	& $\mathbf{80.3}$ & $\mathbf{94.1}$ & $\mathbf{103.3}$ & $\mathbf{101.0}$ & $\mathbf{128.7}$ & $\mathbf{151.3}$ & $\mathbf{166.4}$ \\ 
    \hline
    scenarios  & \multicolumn{4}{c|}{purchases} & \multicolumn{4}{c|}{sitting} & \multicolumn{4}{c|}{sitting down} & \multicolumn{4}{c|}{taking photo} \\
    \hline
    milliseconds & 560ms & 720ms & 880ms & 1000ms & 560ms & 720ms & 880ms & 1000ms& 560ms & 720ms & 880ms & 1000ms & 560ms & 720ms & 880ms & 1000ms  \\ 
    \hline
    \rowcolor{gray}
    Res. Sup. \cite{DBLP:conf/cvpr/MartinezB017}& 122.1 &137.2 &148.0 &154.0 &113.7 &130.5 &144.4& 152.6& 138.8 &159.0& 176.1&187.4 &110.6 &128.9 &143.7& 153.9  \\
    convSeq2Seq \cite{DBLP:conf/cvpr/LiZLL18}&  111.3& 129.1& 143.1 &151.5 &82.4 &98.8 &112.4 &120.7& 106.5 &125.1 &139.8 &150.3 &84.4& 102.4 &117.7& 128.1  \\ \rowcolor{gray}
    LTD \cite{DBLP:conf/iccv/MaoLSL19}& 98.3& 115.1& 130.1& 139.3 &76.4 &93.1 &106.9& 115.7& 96.2& 115.2& 130.8& 142.2& 72.5 &90.9 &105.9& 116.3\\ 
    HRI \cite{DBLP:conf/eccv/MaoLS20} & 95.6& 110.9& 125.0& 134.2& 76.4 &93.1& 107.0& 115.9 & 97.0& 116.1& 132.1& 143.6& 72.1 &90.4 &105.5& 115.9\\ 
    \rowcolor{gray}
    MMA  \cite{DBLP:journals/ijcv/MaoLSL21} &94.5 & 110.2 & 124.4 & 133.1 & 75.8 & 92.3&  106.0 & 115.0 & 96.0 & 115.0 & 130.7& 141.8 & 71.8 & 89.9 & 104.9 & 115.2\\
    siMLPe \cite{DBLP:conf/wacv/GuoDSLAM23} & 93.8&-&-& $\mathbf{132.5}$&75.4&-&-& $\mathbf{114.1}$&95.7 &-&-&142.4 & 71.0&-&-& 112.8\\ 
    \rowcolor{gray}
    Ours & $\mathbf{93.0}$&	$\mathbf{109.1}$&	$\mathbf{123.8}$&	132.7& $\mathbf{74.7}$&	$\mathbf{91.5}$&	$\mathbf{105.4}$&	114.5& $\mathbf{95.1}$&	$\mathbf{114.2}$&	$\mathbf{130.0}$&	$\mathbf{141.6}$&$\mathbf{69.5}$	&$\mathbf{87.2}$&	$\mathbf{101.4}$	&$\mathbf{111.6}$ \\ 
    \hline
    scenarios  & \multicolumn{4}{c|}{waiting} & \multicolumn{4}{c|}{walking dog} & \multicolumn{4}{c|}{walking together} & \multicolumn{4}{c|}{average} \\
    \hline
    milliseconds & 560ms & 720ms & 880ms & 1000ms & 560ms & 720ms & 880ms & 1000ms & 560ms & 720ms & 880ms & 1000ms & 560ms & 720ms & 880ms & 1000ms  \\ 
    \hline
    \rowcolor{gray}
    Res. Sup. \cite{DBLP:conf/cvpr/MartinezB017}& 105.4& 117.3& 128.1 &135.4& 128.7 &141.1& 155.3 &164.5 &80.2 &87.3 &92.8 &98.2 &106.3& 119.4 &130.0 &136.6 \\ 
    convSeq2Seq \cite{DBLP:conf/cvpr/LiZLL18}& 87.3 &100.3 &110.7& 117.7& 122.4& 133.8 &151.1 &162.4 &72.0& 77.7 &82.9 &87.4& 90.7& 104.7& 116.7 &124.2 \\
    \rowcolor{gray}
    LTD \cite{DBLP:conf/iccv/MaoLSL19} & 73.4 &88.2& 99.8&107.5 &109.7 &122.8& 139.0 &150.1 &55.7 &61.3& 66.4 &69.8&78.3 &93.3 &106.0 &114.0 \\
    HRI \cite{DBLP:conf/eccv/MaoLS20} & 74.5& 89.0& 100.3& 108.2 &108.2&120.6 &135.9 &146.9& 52.7 &57.8& 62.0& 64.9 &77.3& 91.8 &104.1& 112.1 \\
    \rowcolor{gray}
    MMA  \cite{DBLP:journals/ijcv/MaoLSL21} &72.7 & 86.9 & 97.6 & 105.1 & $\mathbf{105.1}$ & $\mathbf{117.5}$ & $\mathbf{131.6}$ & 141.4 & 51.2 & 56.2 & 60.3 & 63.2 & 75.9 & 90.4 & 102.5&  110.1\\
    siMLPe \cite{DBLP:conf/wacv/GuoDSLAM23} & 71.6 &-&-& 104.6 &105.6&-&-& $\mathbf{141.2}$ & 50.8&-&-& $\mathbf{61.5}$ &75.7 &90.1&101.8&109.4 \\ 
    \rowcolor{gray}
    Ours & $\mathbf{69.0}$	&$\mathbf{83.8}$	&$\mathbf{95.4}$	&$\mathbf{103.4}$& 112.0&	124.2&	137.7	&147.5 &$\mathbf{48.5}$&	$\mathbf{53.5}$&	$\mathbf{58.5}$&	61.9 &$\mathbf{74.5}$&	$\mathbf{89.2}$	&$\mathbf{101.3	}$&$\mathbf{109.2}$\\ 
    \hline
\end{tabular}}
\end{center}
\caption{Long-term prediction on Human3.6M dataset. MPJPE in millimeter at particular frames (560ms, 720ms, 880ms, 1000ms) are reported. 256 samples are tested for each action. The best results are highlighted in \textbf{bold}.} 
\label{tab:h36_long}
\end{table*}
In \cref{tab:h36_short} and \cref{tab:h36_long}, we show a detailed quantitative performance comparison of some representative state-of-the-art methods for short-term and long-term predictions on the Human3.6M dataset.
\name~outperforms other approaches on most actions and achieves the best average performance.